National Yang Ming Chiao Tung University

# Training Frozen Feature Pyramid DINOv2 for Eyelid Measurements with Infinite Encoding and Orthogonal Regularization

Chun-Hung Chen

## Abstract

Accurate measurement of eyelid parameters such as Margin Reflex Distances (MRD1, MRD2) and Levator Function (LF) is critical in oculoplastic diagnostics but remains limited by manual, inconsistent methods. This study evaluates deep learning models—SE-ResNet, EfficientNet, and the vision transformer-based DINOv2—for automating these measurements using smartphone-acquired images. We assess performance across frozen and fine-tuned settings, using MSE, MAE, and $R^2$ metrics.

DINOv2, pretrained through self-supervised learning, demonstrates superior scalability and robustness, especially under frozen conditions ideal for mobile deployment. Lightweight regressors such as MLP and Deep Ensemble offer high precision with minimal computational overhead. To address class imbalance and improve generalization, we integrate focal loss, orthogonal regularization, and binary encoding strategies.

Our results show that DINOv2 combined with these enhancements delivers consistent, accurate predictions across all tasks, making it a strong candidate for real-world, mobile-friendly clinical applications. This work highlights the potential of foundation models in advancing AI-powered ophthalmic care.

## Table of Content

# Introduction

Accurate eyelid measurements—specifically margin reflex distances (MRD1 and MRD2) and levator function (LF)—are essential for diagnosing and managing eyelid ptosis in oculoplastic surgery. These measurements influence both functional and cosmetic outcomes, but they are still commonly obtained through manual observation or caliper-based methods that are subjective, inconsistent, and difficult to standardize. As mobile diagnostics and remote care expand, there is a growing need for reliable, automated measurement systems that can function effectively across clinical and non-clinical settings.

Advances in deep learning have made it possible to address this need. CNN-based architectures such as SE-ResNet [1] and EfficientNet [2] have demonstrated improved accuracy and reproducibility in smartphone-based eyelid measurements [3] However, these models typically require large amounts of annotated data and extensive retraining to adapt across domains—barriers that limit scalability and practical deployment, especially on mobile platforms.

To overcome these limitations, this study investigates DINOv2 [4], a vision foundation model built on self-supervised learning. By learning representations directly from unlabeled data, DINOv2 addresses the bottleneck of annotated dataset availability, a persistent issue in medical imaging. Incorporating recent architectural advances in vision transformers [5], DINOv2 achieves robust feature extraction and exhibits strong transferability even under frozen configurations—qualities that align well with the efficiency and scalability required for mobile health applications.

We evaluate DINOv2 against SE-ResNet and EfficientNet across multiple regression metrics (MSE, MAE, and $R^2$), testing both pretrained and fine-tuned settings. In doing so, we examine model performance, learning dynamics, and deployment feasibility under real-world constraints. By integrating clinical needs with cutting-edge model design, this study aims to demonstrate how foundation models like DINOv2 can support accurate, accessible, and scalable eyelid analysis in ophthalmology.

# Literature Review

### Clinical Importance of Eyelid Measurement

In oculoplastic surgery, the accuracy of eyelid measurements—particularly margin reflex distances (MRD1 and MRD2) and levator function (LF)—is critical for diagnosing and treating eyelid ptosis, a condition that affects both visual function and aesthetic appearance. These measurements inform surgical planning and directly impact patient outcomes. However, traditional assessment methods rely on manual observation or caliper-based techniques, which are prone to variability and subjectivity. Such inconsistencies can result in suboptimal surgical decisions, especially in borderline or complex cases. As a result, the development of objective, reproducible, and automated measurement methods has become a central focus in clinical research. The need for precision is not only practical but also essential to improving consistency across surgeons, enhancing preoperative evaluation, and ensuring equitable care—especially as telemedicine and mobile diagnostics become increasingly prevalent in ophthalmic practice [6], [7]

### Evolution of Computer Vision in Medical Imaging

The field of computer vision has evolved significantly, particularly in its application to medical imaging, where accurate analysis of complex visual data is paramount. Early models, such as ResNet, introduced the concept of residual learning to address the vanishing gradient problem, enabling the training of much deeper networks. SE-ResNet [1], an extension of ResNet, incorporated Squeeze-and-Excitation blocks that adaptively recalibrate feature responses, improving the model's representational capacity. Similarly, EfficientNet [2] focused on optimizing network depth, width, and resolution, resulting in more efficient and accurate models that are well-suited to real-time medical image analysis.

However, recent advancements have led to a paradigm shift with the introduction of vision transformers, which use self-attention mechanisms to process images as sequences of patches. These models capture global dependencies within data, offering substantial improvements in tasks requiring detailed analysis, such as medical image segmentation and classification. Vision transformers, particularly with their ability to learn intricate spatial and structural patterns, represent a significant departure from convolutional neural networks (CNNs). These innovations have opened the door to more flexible, scalable, and efficient methods for complex tasks in medical imaging, such as assessing eyelid features in oculoplastic surgery.

**Breakthroughs in Self-Supervised Learning**

Self-supervised learning (SSL) has emerged as a solution to the limited availability of labeled medical data. SSL models learn directly from data structure, reducing dependence on manual annotations [8] Early SSL approaches like SimCLR and BYOL used contrastive learning to align representations of augmented images, while DINO [9] introduced a self-distillation mechanism that taught the model to align its own augmented views.

DINOv2 [4], [5] builds on these foundations with a more scalable and stable transformer-based framework. Its strong performance on vision tasks—even without fine-tuning—makes it particularly well-suited for clinical imaging scenarios where data is limited, but accuracy is critical. Unlike CNNs, which often require full retraining for new tasks, DINOv2 supports zero-shot and few-shot applications with minimal overhead. This efficiency makes it viable for mobile deployment, where computational constraints and data variability are common.

While DINOv2 generates robust features, the way these are structured and extracted is equally critical—especially in tasks that involve spatial reasoning, such as measuring eyelid distances. This leads to the importance of hierarchical feature engineering.

**Hierarchical Feature Engineering**

In medical image tasks where spatial precision is essential, hierarchical feature extraction enables models to capture both coarse global context and fine local detail. U-Net [10], a widely used architecture in biomedical segmentation, combines downsampled encodings with upsampled decodings through skip connections. This structure preserves spatial detail while incorporating semantic depth, improving delineation of fine structures like eyelid creases or margins.

Similarly, Feature Pyramid Networks (FPNs) [11] create multi-scale feature hierarchies by fusing high-level semantics with low-level spatial resolution. Originally developed for object detection,

FPNs have proven valuable in medical tasks involving small or variable-sized targets. For eyelid measurement, these architectures support accurate landmark localization and enhance stability across image conditions, including variable lighting, pose, and device type.

**Recent Development of Regressor Head**

The rise of attention-based architectures has significantly improved regression performance on complex inputs like medical images. Transformers [12], with their global attention mechanisms, can model cross-feature relationships more effectively than traditional MLPs or CNNs. This allows better handling of structural dependencies, which is especially beneficial in predicting continuous anatomical parameters.

The iTransformer [13] refines this further by inverting the standard attention pipeline for better scalability and efficiency—traits beneficial for mobile applications. Meanwhile, BERT [14], though initially developed for text, has shown versatility in regression when adapted to structured or spatial data. Models like FT-Transformer [15] and TabTransformer [16] extend transformer capabilities to tabular datasets by embedding numerical and categorical features, offering improved performance in domains like health informatics.

In parallel, Deep Ensembles [16] have become a popular strategy for uncertainty-aware regression. By aggregating outputs from multiple independently trained models, ensembles improve robustness and predictive confidence—important in clinical contexts where wrong predictions may have serious implications.

**Long-Tailed Learning and Orthogonal Regularization**

Eyelid measurement tasks, like many clinical problems, are often affected by long-tailed data distributions, where rare but clinically important cases are underrepresented. This imbalance, analogous to the foreground-background problem in object detection, leads to biased models that perform poorly on extreme values. Focal Loss [18] mitigates this by emphasizing harder examples and down-weighting well-classified ones, directing the model's attention to minority instances that may hold greater diagnostic weight.

Generalization remains another critical challenge, especially when deploying models across diverse patient populations or imaging conditions. One effective approach is orthogonality regularization, which encourages flatter minima by constraining weight matrices to be orthogonal —an optimization geometry associated with better generalization [19] Orthogonality has been shown to improve feature diversity and stability in convolutional models [20], and to promote disentangled latent structures [21]

In NLP, orthogonal techniques like Orthogonal Subspace Learning [22] and Orthogonal Weight Modification (OWM) [23] have supported continual learning by reducing task interference. These principles are transferable to clinical settings, where AI models must adapt to evolving datasets without forgetting prior knowledge. Together, focal loss and orthogonal regularization provide complementary mechanisms to improve both the fairness and robustness of models deployed in real-world healthcare environments.

# Methodology

## Data Preparation

The study's data preparation phase closely aligns with the methodology employed in Chen et al.'s research. A total of 822 eyes from 411 volunteers were photographed using a smartphone, capturing various gaze positions. These photographs were standardized using a consistent scale and processed to ensure uniformity. The photographs underwent meticulous normalization, a process vital for maintaining consistency in input data for deep learning models. This normalization involved segmenting and aligning the photographs to focus on relevant eye regions for accurate measurement of Margin Reflex Distance (MRD1 and MRD2) and Levator Function (LF).

The dataset was divided into training, validation, and testing groups, with 90% of the data allocated to the training/validation group and 10% reserved for testing. Within the training/validation group, an 80-20 split was further implemented for training and validation purposes. This division ensures a comprehensive evaluation of the model's performance across different subsets of the data.

| Model | | Cases, n (%) | Males, n (%) |
|---|---|---|---|
| MRD1 | Total | 822 (100.0) | 154 (18.7) |
| | Training group | 740 (90.0) | 142 (19.2) |
| | Testing group | 82 (10.0) | 12 (14.6) |
| MRD2 | Total | 822 (100.0) | 154 (18.7) |
| | Training group | 740 (90.0) | 142 (19.2) |
| | Testing group | 82 (10.0) | 12 (14.6) |
| LF | Total | 685 (100.0) | 122 (17.8) |
| | Training group | 617 (90.0) | 113 (8.3) |
| | Testing group | 68 (13.2) | 9 (13.2) |

Table 1. Case numbers and sex ratios in each model.

| Measurements | N | Mean (SD) | Range |
|---|---|---|---|
| MRD1 | 822 | 2.59 (1.21) | 0.00-6.00 |
| MRD2 | 822 | 5.51 (0.83) | 1.50-10.00 |
| LF | 685 | 12.1 (2.12) | 3.50-18.00 |

Table 2. Summary of gold-standard measurements.

| Metric | MRD1 | MRD2 | LF |
|---|---|---|---|
| MAE | 0.007 | 0.008 | 0.018 |
| MSE | 0.005 | 0.001 | 0.002 |

Table 3. Reliability of gold-standard measurements (actual values) manually performed by the two doctors.

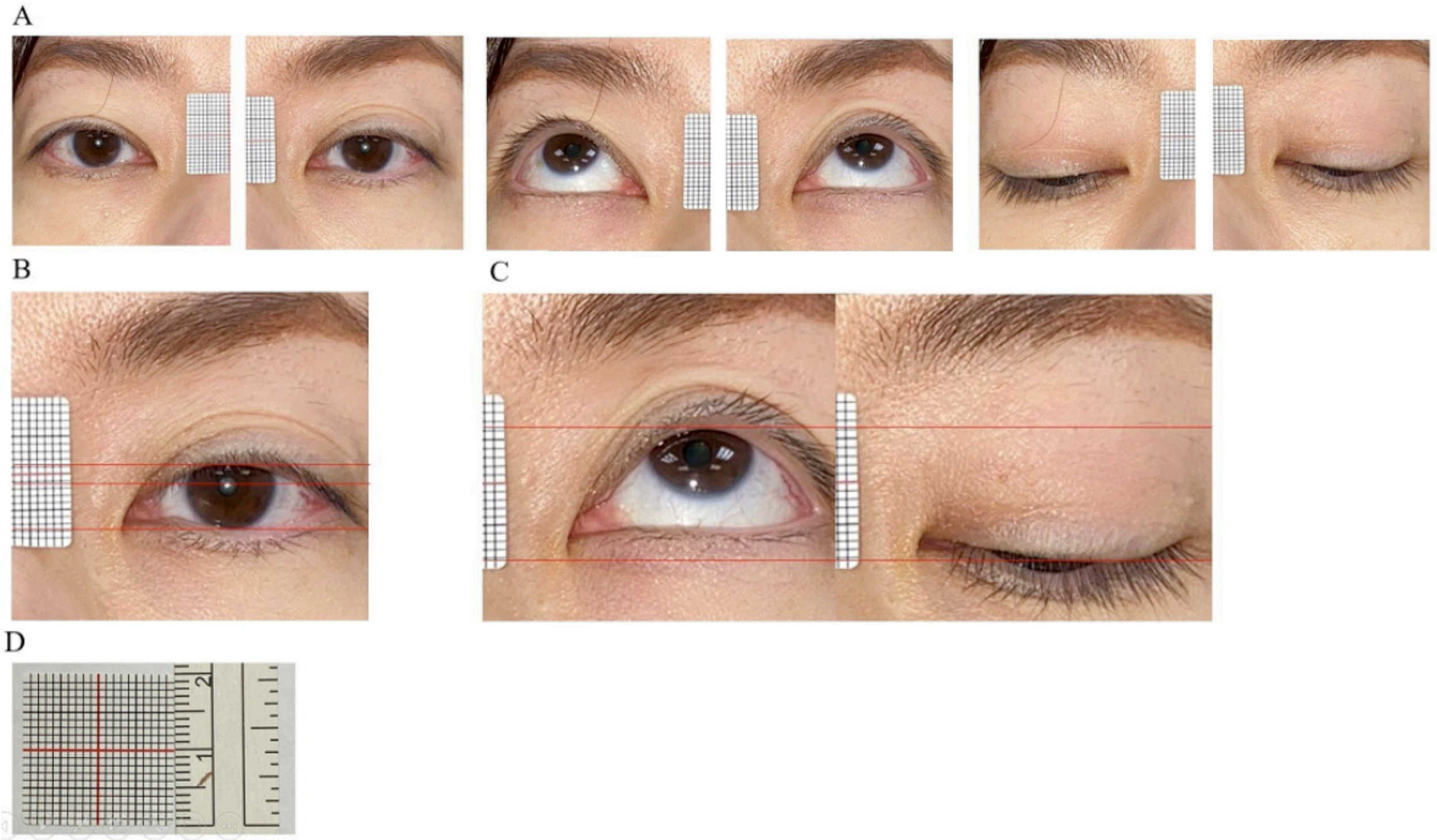

Figure 1. Photographs and gold-standard measurements (A) Six orbital photographs, including bilateral primary gaze, up-gaze, and down-gaze, were taken by a smartphone. (B) The primary gaze photograph was then magnified for MRD1 and MRD2 measurements. (C) The up-gaze and down-gaze photographs were then magnified for LF measurements. (D) A 20×20-mm scale.

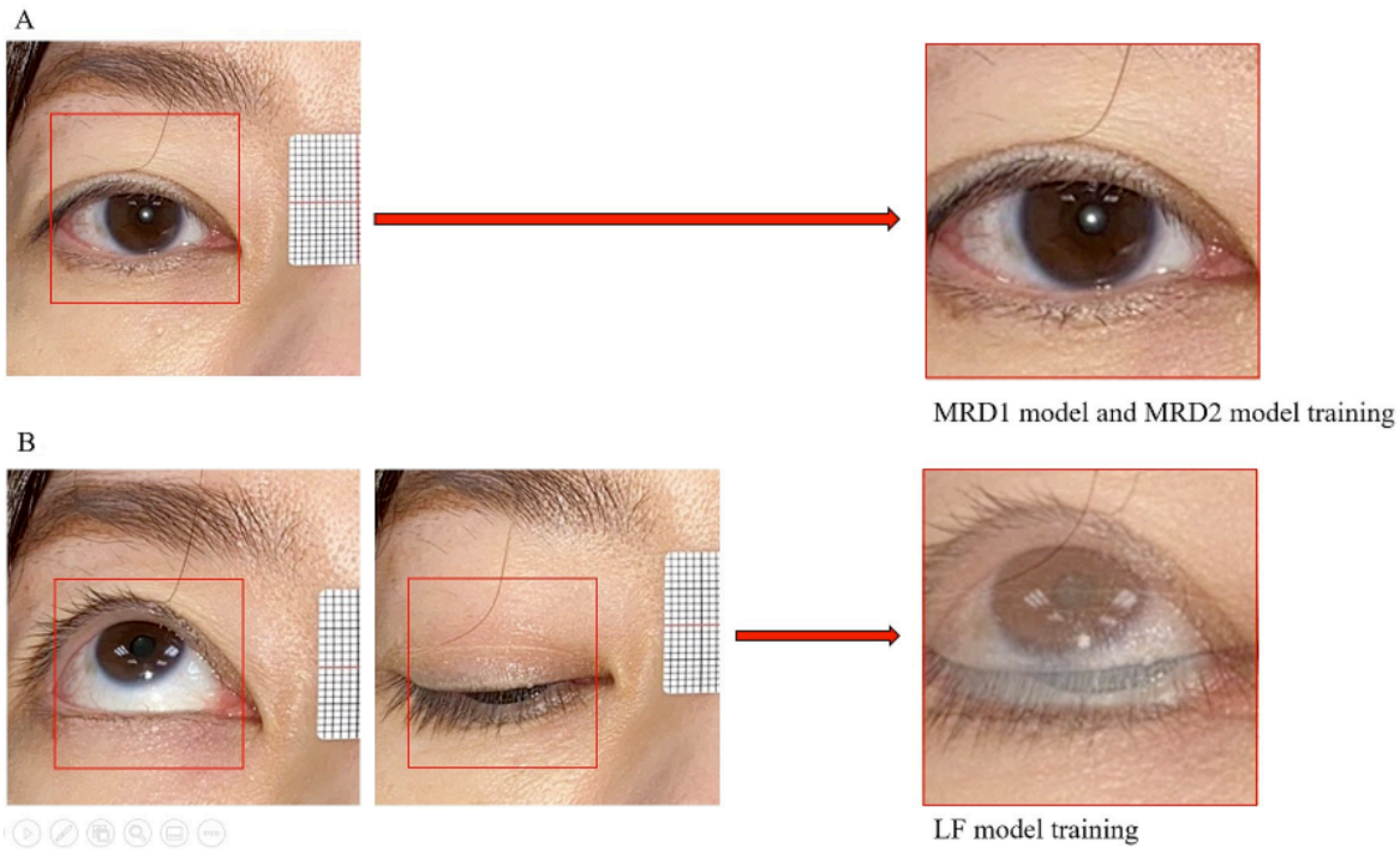

Figure 2. Photograph Normalization. (A) Autosegmentation of primary-gaze orbital photographs. These photographs are considered the “normalized eye photographs” for MRD1 and MRD2 model training. (B) Autosegmentation of up- and down-gaze orbital photographs, which were then merged into one photograph. These photographs are considered the “normalized eye photographs” for LF model training.

**Model Architecture**

The study compares three different deep learning architectures: SE-ResNet, EfficientNet, and DINOv2 (ViT-base). SE-ResNet, known for its ability to focus on relevant features within an image, and EfficientNet, recognized for its efficiency and scalability, have both demonstrated commendable performance in image analysis tasks. DINOv2, the latest addition, is a vision transformer model that distinguishes itself with its self-supervised learning approach. This model is pre-trained on large datasets, allowing it to capture intricate patterns and details in visual data.

All three models utilize a similar strategy where their extracted embeddings are fed into the Multi-Layer Perceptron (MLP), Attention, Transformer, iTransformer, Bert, FTTransformer, TabTransformer, and Deep Ensemble for the regression tasks associated with MRD1, MRD2, and LF measurements. This approach leverages the strengths of each model in feature extraction and harnesses the power of MLPs for precise regression analysis.

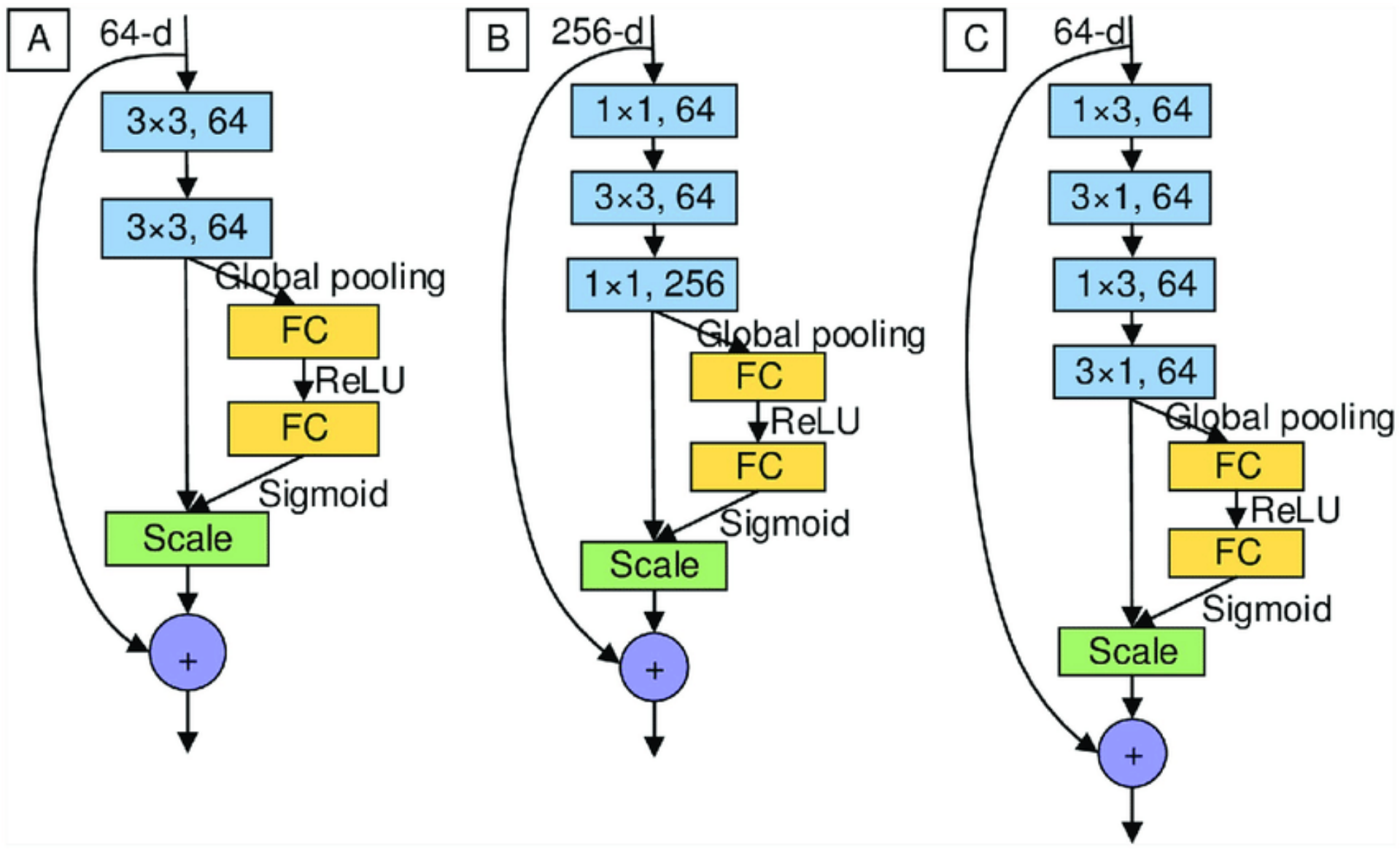

Figure 3. The SE-ResNet Module Architecture. (A) Basic SE-ResNet module. (B) Bottleneck SE-ResNet module. (C) Small SE-ResNet module.

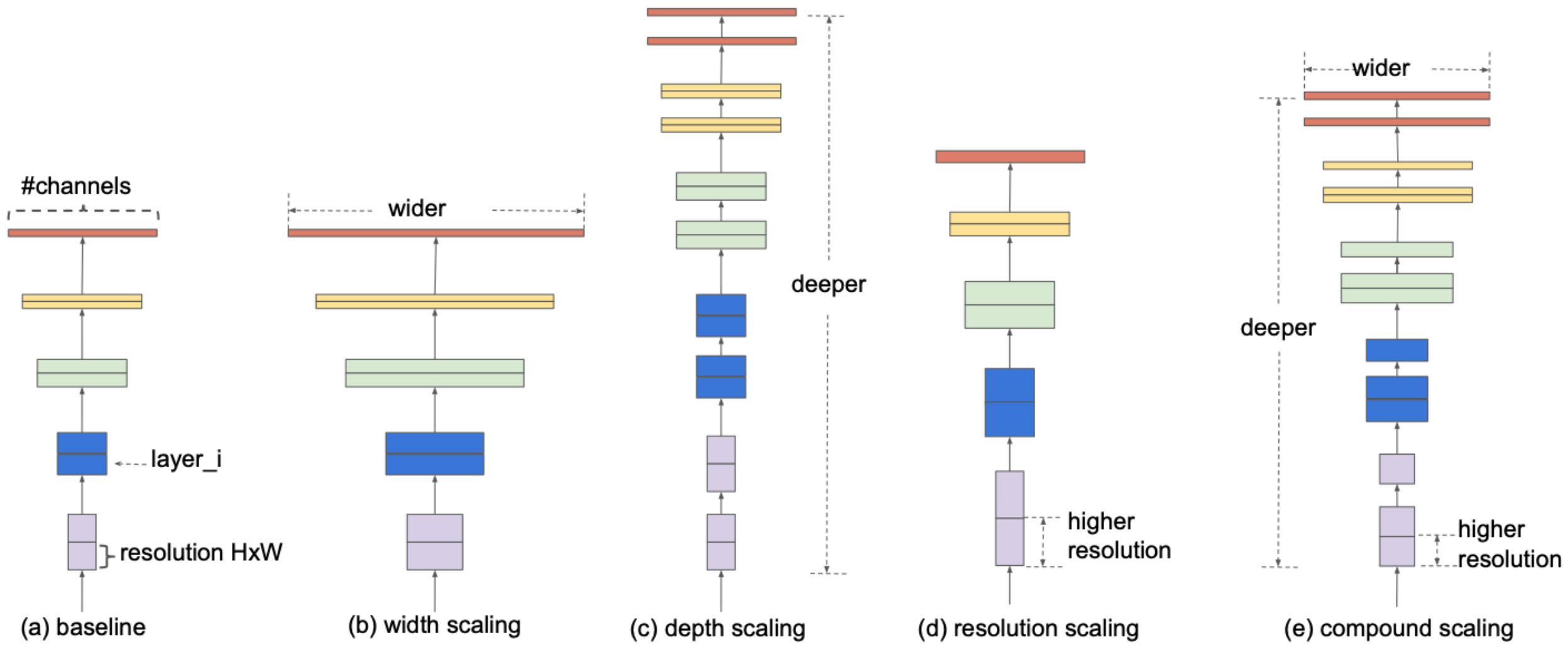

Figure 4. The EfficientNet Model Scaling. (a) is a baseline network example; (b)-(d) are conventional scaling that only increases one dimension of network width, depth, or resolution. (e) is our proposed compound scaling method that uniformly scales all three dimensions with a fixed ratio.

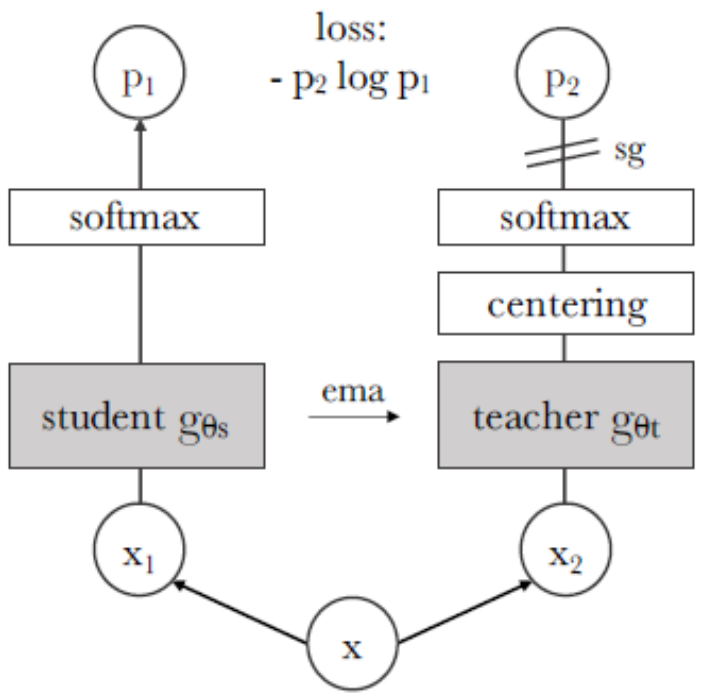

**Algorithm 1** DINO PyTorch pseudocode w/o multi-crop.

```
# gs, gt: student and teacher networks
# C: center (K)
# tps, tpt: student and teacher temperatures
# l, m: network and center momentum rates
gt.params = gs.params
for x in loader: # load a minibatch x with n samples
    x1, x2 = augment(x), augment(x) # random views

    s1, s2 = gs(x1), gs(x2) # student output n-by-K
    t1, t2 = gt(x1), gt(x2) # teacher output n-by-K

    loss = H(t1, s2)/2 + H(t2, s1)/2
    loss.backward() # back-propagate

    # student, teacher and center updates
    update(gs) # SGD
    gt.params = l*gt.params + (1-l)*gs.params
    C = m*C + (1-m)*cat([t1, t2]).mean(dim=0)

def H(t, s):
    t = t.detach() # stop gradient
    s = softmax(s / tps, dim=1)
    t = softmax((t - C) / tpt, dim=1) # center + sharpen
    return - (t * log(s)).sum(dim=1).mean()
```

Figure 5. The DINO Framework. The student and teacher networks, sharing the same architecture but with distinct parameters, each receive a uniquely transformed version of an input image. The teacher's output is batch-centered, and both networks produce a K-dimensional feature normalized by temperature softmax. Their outputs are compared using cross-entropy loss. A stop-gradient operator on the teacher ensures only the student network learns, while the teacher's parameters are gradually updated to reflect the student's, using an exponential moving average method.

**Feature Pyramid Architecture and Data Imbalance**

To address the need for capturing multi-scale information in medical images, the study incorporates Feature Pyramid Networks (FPNs) [11] into the model pipeline. FPNs are particularly effective for tasks involving detection or measurement of small, localized structures—such as eyelid landmarks—by combining high-resolution spatial information with high-level semantic features. In this study, FPNs are applied on top of extracted embeddings from SE-ResNet, EfficientNet, and DINOv2 to enhance the regression performance for MRD1, MRD2, and LF prediction. The FPN layers allow the models to more effectively encode both global facial geometry and fine periocular detail.

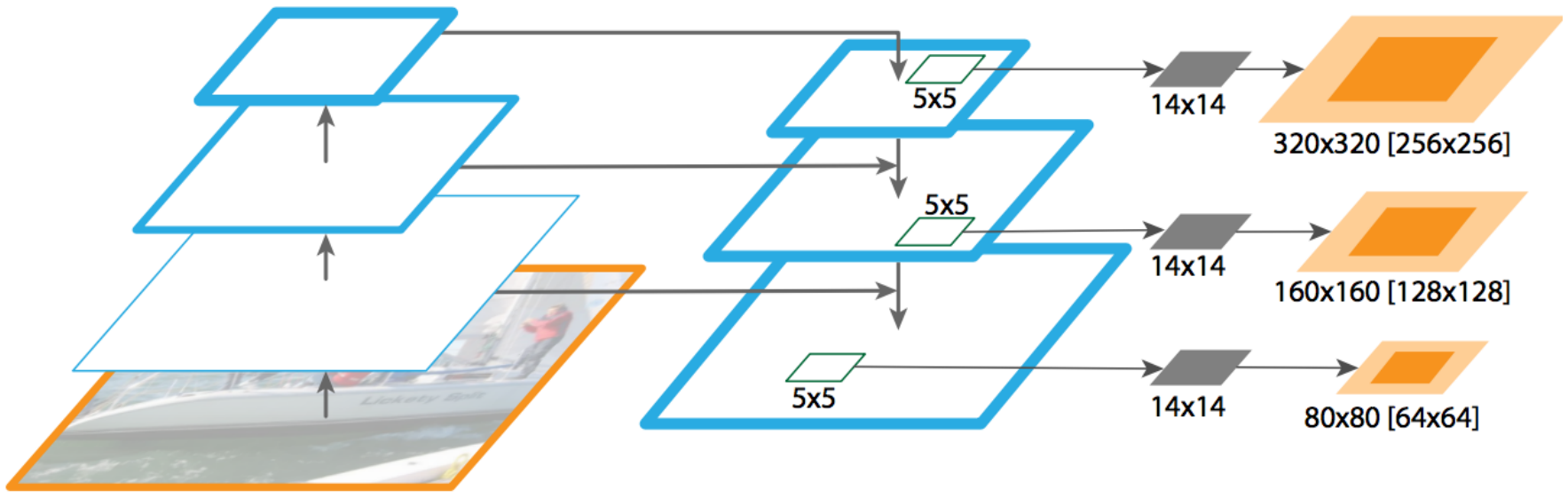

Figure 6. Feature Pyramid Network (FPN) architecture. Multi-scale feature maps are generated by merging high-level semantic information from deeper layers with spatially rich representations from earlier layers. Each level undergoes a 5×5 convolution, followed by upsampling and fusion, producing uniform 14×14 spatial features at multiple input resolutions (320×320, 160×160, 80×80). This allows the model to detect and regress across varied spatial scales, improving performance on small or fine-grained anatomical targets.

However, given the clinical variability in eyelid anatomy and the natural imbalance in measurement distributions (e.g., ptosis cases vs. normal eyelids), the dataset presents a long-tailed problem. Standard loss functions tend to favor common cases, potentially ignoring underrepresented but clinically critical patterns. To address this, the study applies Focal Loss[18], which down-weights easy examples and focuses training on harder, mispredicted samples. This adjustment enhances sensitivity to rare presentations and stabilizes learning across the full spectrum of values, improving overall regression reliability.

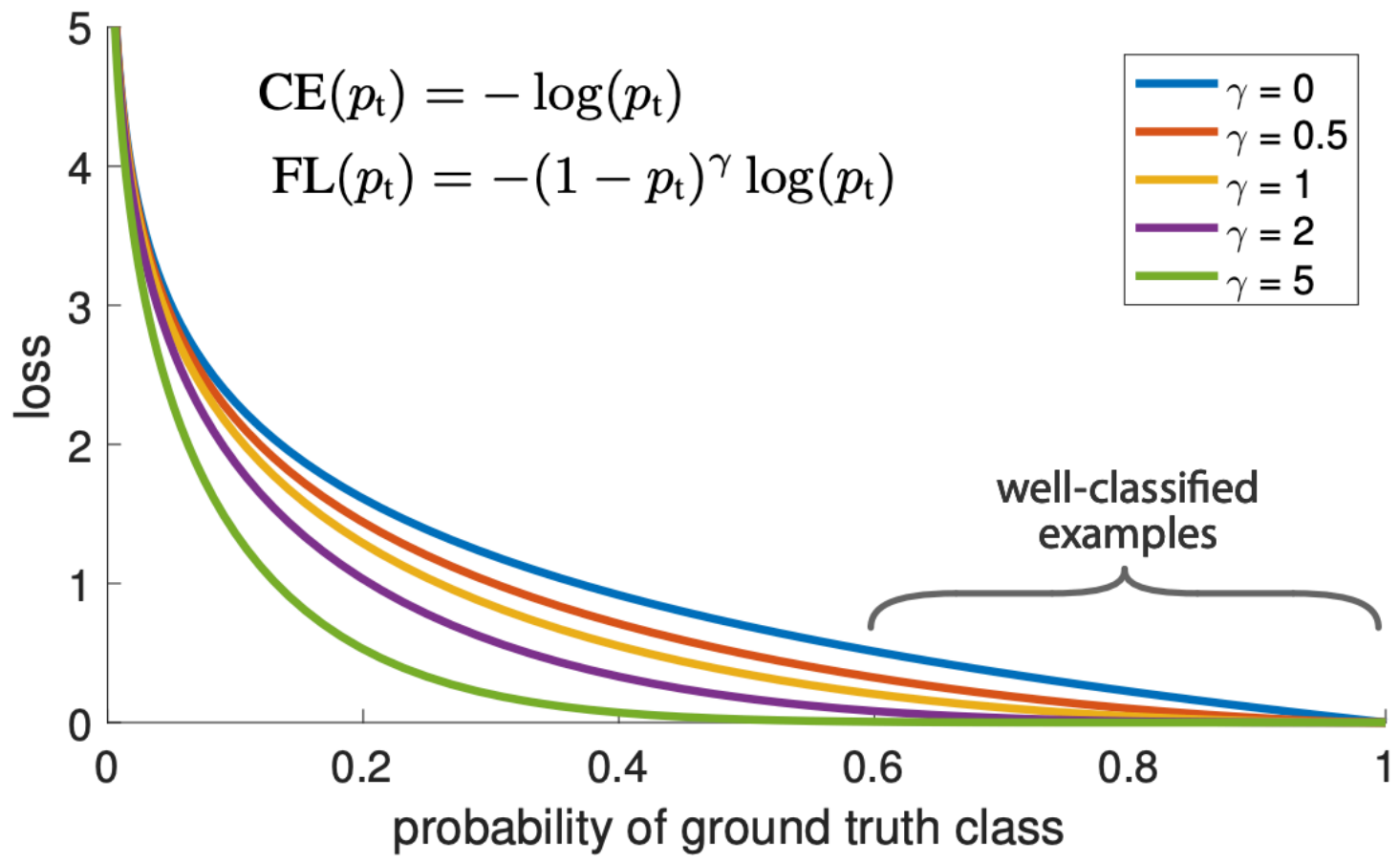

Figure 7. Visualization of the Focal Loss function compared to standard Cross Entropy (CE). As the focusing parameter $\gamma$ increases, the loss contribution from well-classified examples (high

predicted probability for the ground truth class) is reduced. This reweighting mechanism enables the model to focus more on hard, misclassified examples—helpful in addressing class imbalance in dense prediction tasks such as eyelid measurement with skewed data distributions.

**Domain Specific Self-Supervised Pretraining**

To enhance feature extraction in a clinically specific domain, this study leverages domain-specific self-supervised pretraining using the DINOv2 framework. Although DINOv2 is pretrained on large-scale, diverse datasets, general-purpose visual features may not fully capture the subtle anatomical nuances necessary for eyelid measurement tasks. To address this, we adapted DINOv2 to a domain-specific context by applying it to a large corpus of unlabeled periocular and orbital images collected under clinical protocols.

The self-supervised objective is to learn meaningful and transferable representations from unlabeled ophthalmic images without human annotations. As illustrated in Figure 8, the framework follows a student-teacher paradigm. Both networks receive differently augmented views of the same image; the teacher's output, processed through temperature-scaled softmax and centered by batch normalization, serves as a stable reference for the student. A stop-gradient mechanism ensures that only the student updates during training, while the teacher is updated via exponential moving average of the student weights. This distillation strategy allows the model to capture rich, invariant visual representations unique to eyelid structures—such as crease lines, lid thickness, and levator excursion patterns.

Once pretrained, the domain-adapted backbone is frozen and paired with lightweight regression heads (MLP, Transformer-based, or ensemble variants) for downstream MRD1, MRD2, and LF prediction tasks. This setup enables efficient transfer to low-data settings while preserving the specificity required for clinical precision. Domain-specific pretraining not only bridges the gap between general vision features and clinical relevance but also improves performance on difficult edge cases often underrepresented in labeled datasets.

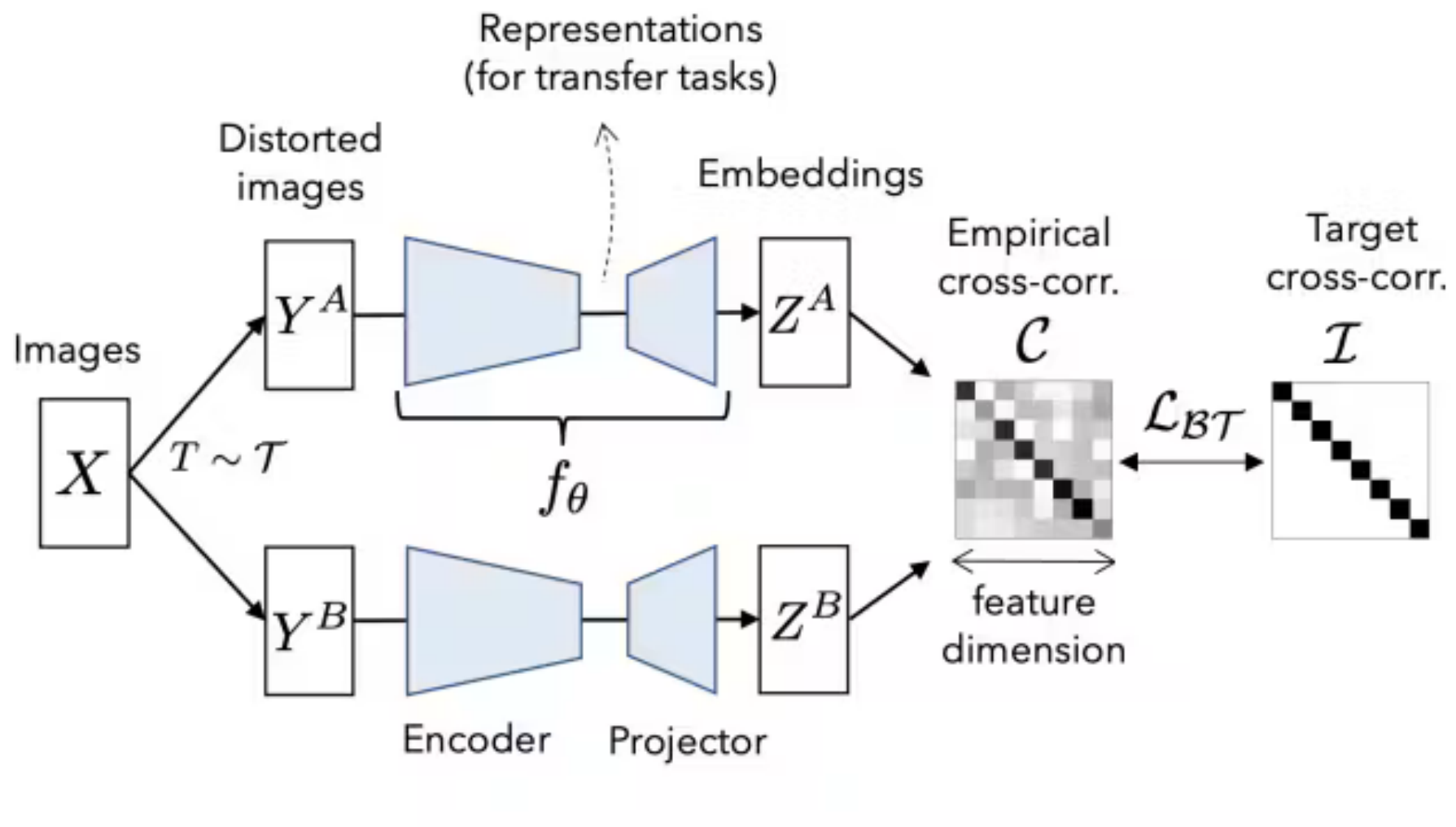

Figure 8. Illustration of the domain-specific self-supervised pretraining framework using DINOv2. Two augmented views of an unlabeled eyelid image are passed through a student and teacher network with shared architecture. The student learns to align its output with the teacher's, which is updated via exponential moving average. This process enables learning of ophthalmology-specific visual features without labels, improving downstream regression performance.

**Orthogonality of Model Parameters**

To improve model generalization and mitigate overfitting—especially in tasks with limited or imbalanced clinical data—the study incorporates orthogonality regularization into the training process. This approach constrains weight updates to remain orthogonal to the subspace of previously learned information, thereby encouraging diverse and non-redundant representations across layers.

Building on principles from Orthogonal Weight Modification (OWM) and related works in continual learning [OWM, Orthogonal Subspace], the method ensures that the learning of new parameters does not interfere destructively with previously learned tasks. As illustrated in Figure 9a, standard gradient updates ($\Delta W^{BP}$) may drift into directions that overwrite existing knowledge, whereas orthogonal updates ($\Delta W^{OWM}$) project the gradient onto a safe subspace, preserving previously acquired information.

This concept extends naturally to multi-task and clinical regression settings. For example, when fine-tuning models for MRD1, MRD2, and LF on overlapping image features, orthogonality constraints help the model retain distinct, non-conflicting parameter sets for each task. As shown in Figure 9b, while traditional SGD may result in task interference, OWM enables parameter updates to remain within safe zones of the parameter space, maintaining performance across all tasks.

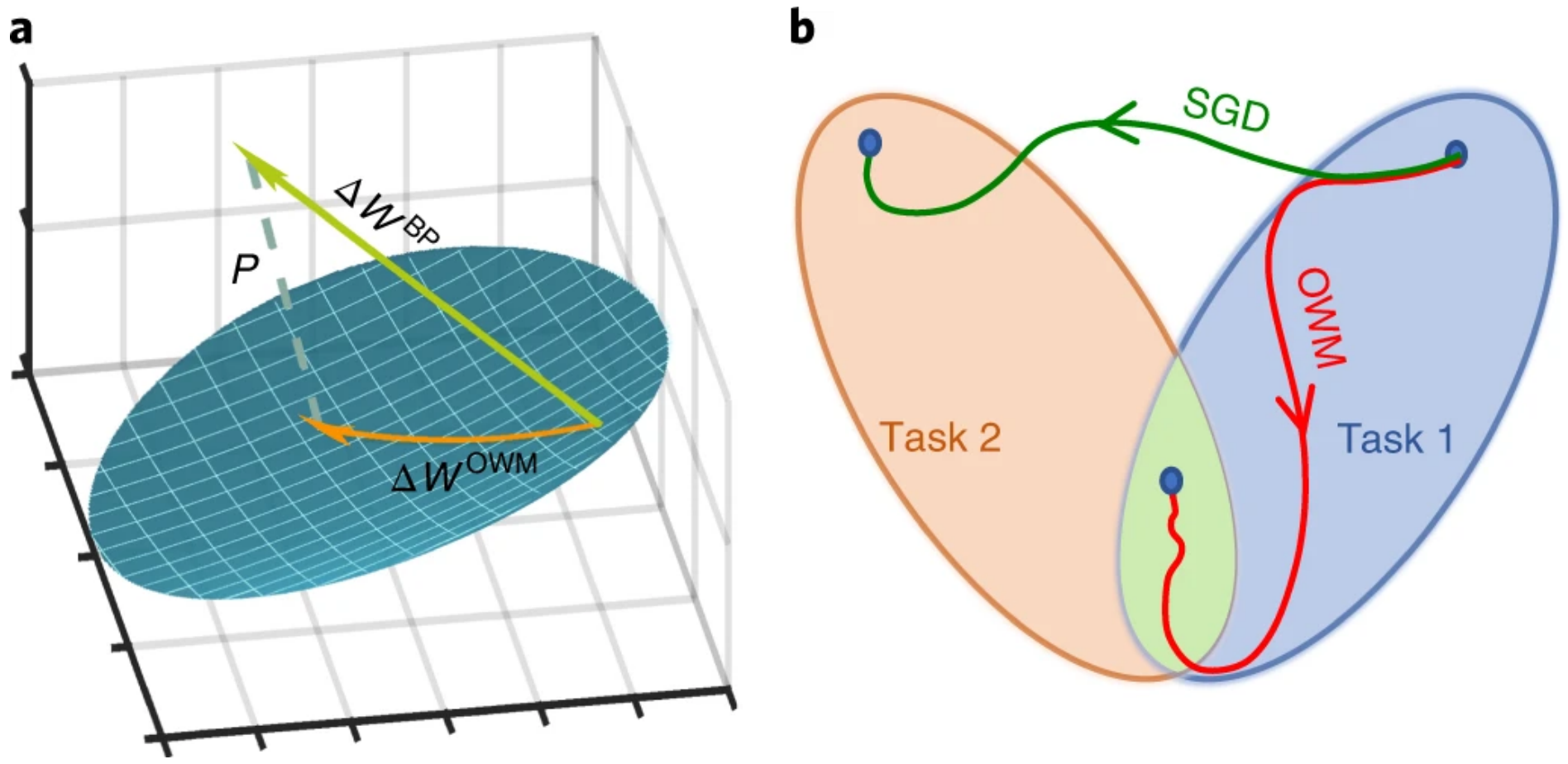

Figure 9. Illustration of orthogonal weight modification (OWM) in continual learning. (a) Gradient updates (ΔW) are projected onto the orthogonal subspace (P) to avoid interference with previously learned knowledge. (b) Compared to standard stochastic gradient descent (SGD), OWM preserves performance on prior tasks (Task 1) while learning new tasks (Task 2) by constraining updates within a non-overlapping parameter space. This promotes generalization and prevents catastrophic forgetting, making it suitable for lifelong learning in clinical AI systems.

**Infinite Encoding and Regression Precision**

To address the limitations of traditional regression models—particularly in handling data imbalance and ensuring fine-grained precision—we propose a novel formulation that transforms regression into a binary classification problem using bitwise encoding. By discretizing the continuous regression target into a series of binary encodings, each representing a finer-grained subdivision of the value range, we achieve theoretically infinite resolution through progressively deeper encodings.

As shown in Figure 10, the regression range is quantized via bit-level representations:

- A 1-bit encoding yields two classes (e.g., low vs. high),
- A 2-bit encoding maps into four discrete bins,
- A 3-bit encoding extends this to eight finer subdivisions,
- And so on, up to ∞-bit encoding, which approximates the continuous domain with arbitrarily high precision.

Each additional bit doubles the number of representable values, effectively increasing the regression resolution. Importantly, this approach recasts regression as multi-label classification, enabling the application of robust classification-specific techniques, including long-tailed learning, focal loss, and balanced sampling, all of which are better established and empirically validated in the classification domain.

This reformulation not only improves performance on sparse or imbalanced regression targets but also facilitates calibrated uncertainty estimation through binary entropy aggregation across bits—providing clinicians with both accurate and interpretable outputs in tasks like MRD1, MRD2, and LF estimation.

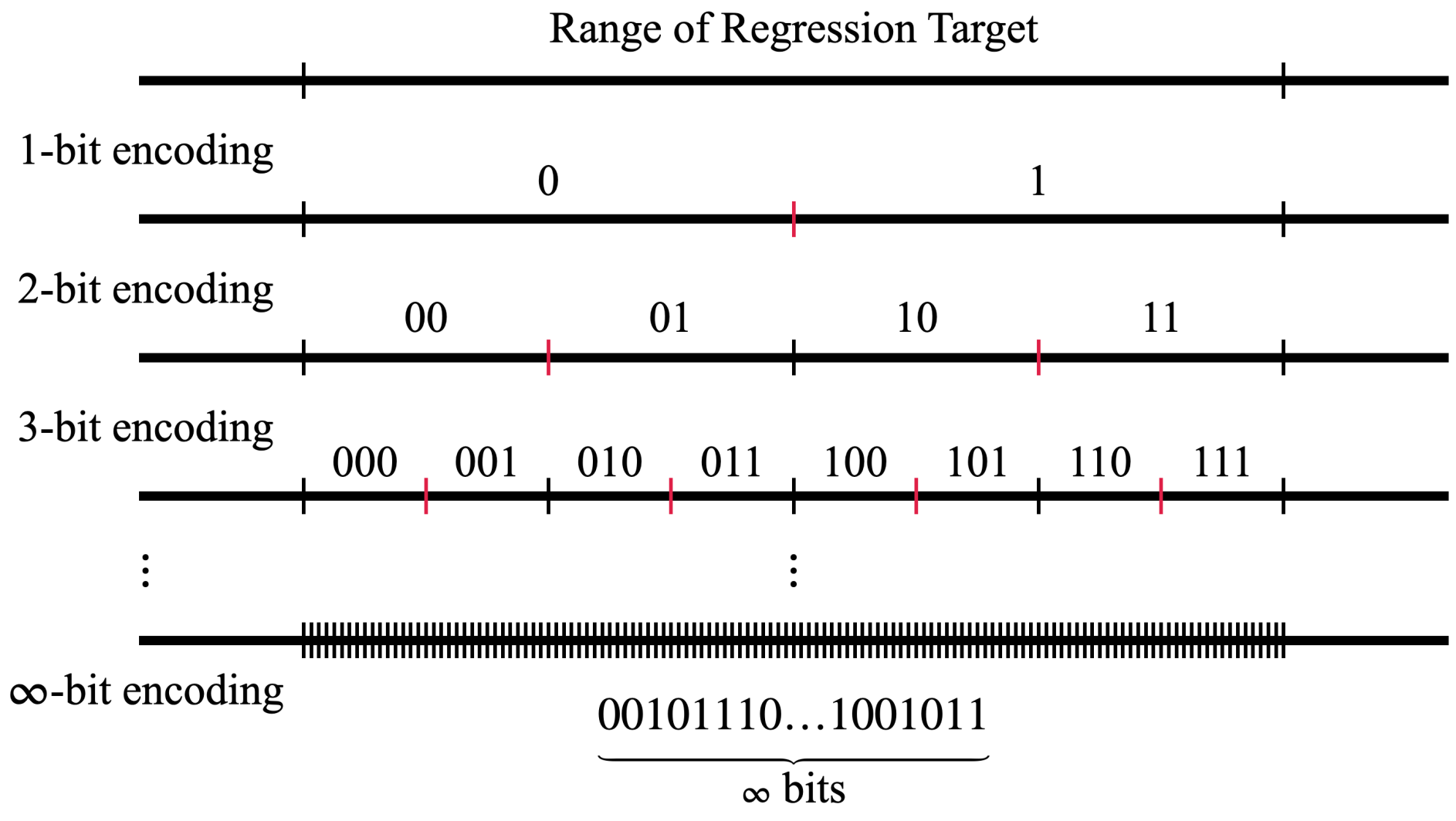

Figure 10. Each bit increases the granularity of the encoded regression range. A 1-bit representation creates two partitions, 2-bit yields four, and so on. With ∞-bit encoding, the regression space becomes densely quantized, enabling near-continuous representation. This hierarchical structure converts continuous-valued regression into a multi-bit classification problem, allowing the use of long-tailed classification strategies and enabling fine-grained, scalable prediction.

**Implementation Details**

The implementation of these models adhered to specific parameters to ensure optimal performance and accurate results. The loss function used was Mean Squared Error (MSE), and the Adam optimizer was chosen for its efficiency, with a learning rate set at 1e-3. Considering the complexity of the models and the size of the data, a batch size of 4 was selected. Each model underwent training for 20 epochs, with an option to freeze the backbone model to evaluate the impact of utilizing pre-trained weights versus training from scratch.

During training, the weight, activity log, and learning curve of each condition were meticulously recorded. This not only provides transparency in the model's learning process but also aids in identifying areas for potential improvement. The entire implementation was developed predominantly using PyTorch and trained on a high-performance Nvidia RTX 6000 Ada graphics processing unit, ensuring efficient computation and processing capabilities.

**Evaluation Matrix**

The evaluation of the model’s performance was carried out using a set of well-established metrics, including Mean Squared Error (MSE), Mean Absolute Error (MAE), and the R2 score. These metrics provide a comprehensive understanding of the model’s accuracy, precision, and overall predictive power. The learning curve, an essential tool for visualizing the models' learning progression over epochs, offers insights into their training dynamics, including aspects such as convergence and overfitting.

# Results and Discussion

## Enhanced Performance of Frozen DINOv2

Figure 11 presents a detailed Mean Squared Error (MSE) comparison among various backbone architectures for predicting margin reflex distances (MRD1 and MRD2) and levator function (LF). The bar graph delineates each model's performance under two scenarios: with a frozen backbone (indicated in blue) and without freezing (shown in green). This distinction helps assess the models' adaptability to new data sets. Specifically, the EfficientNet models show mixed MSE results for MRD1 and MRD2 predictions, where freezing the backbone typically leads to higher error rates. In contrast, the DINOv2 models demonstrate lower MSE, especially when the backbone remains unfrozen, indicating superior adaptation to specific tasks. However, for LF prediction, the frozen condition of DINOv2 results in a marginally increased MSE, suggesting a risk of overfitting or reduced generalization without fine-tuning task-specific data. These observations underscore the significance of model flexibility and the advantages of allowing pre-trained models to adapt to specific dataset nuances in medical imaging.

Additionally, the performance of the DINOv2 model, particularly in its frozen state, highlights its aptness for mobile computing in clinical environments. The model's efficacy in providing accurate and reliable eyelid measurements on smartphones greatly enhances the accessibility and convenience of oculoplastic diagnostics. This capability meets the growing need for mobile health solutions that deliver high-quality performance without excessive computational demands, showcasing the practical benefits of integrating advanced AI models into routine clinical practices.

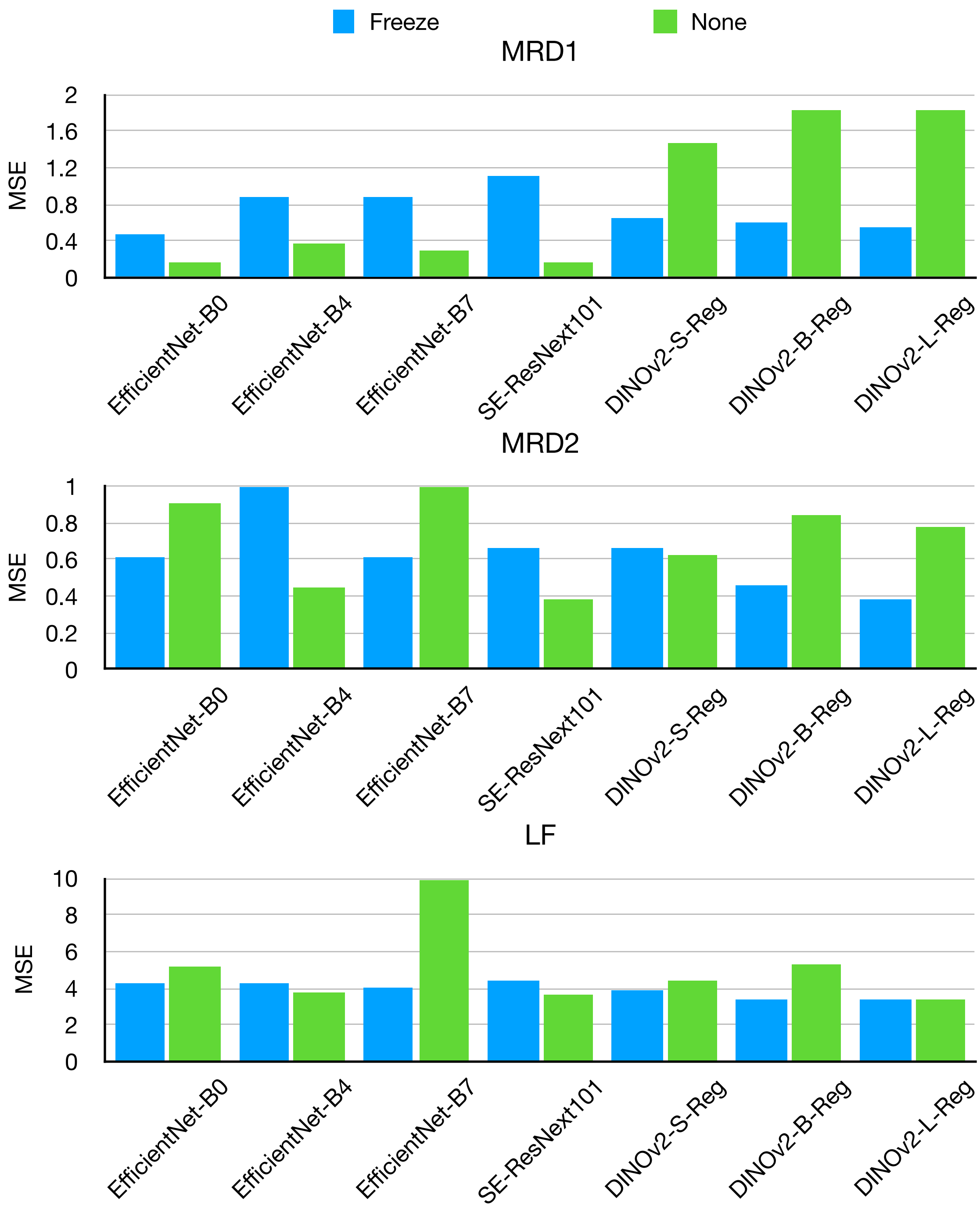

Figure 11. Evaluation of Model Performance in Eyelid Measurement Predictions: A Comparison of Frozen and Unfrozen Backbone Architectures.

**Efficacy of Self-Supervised Pre-training**

In this study, we assess self-supervised pre-training's impact on various Vision Transformer (ViT) models, from small (ViT-S) to large (ViT-L), focusing on eyelid measurement accuracy. The findings, detailed in Table 4, reveal that DINOv2 pre-training notably improves performance in mean squared error (MSE), mean absolute error (MAE), and R2 for MRD1, MRD2, and LF, with larger models showing more significant enhancements. This demonstrates that DINOv2 pre-trained models excel in medical tasks, even under computational limitations, underscoring self-supervised learning's potential in medical imaging.

Furthermore, applying these models in mobile healthcare is highly beneficial. Incorporating DINOv2 pre-training into mobile apps aligns with smartphones' computational limits while improving point-of-care diagnostic quality. This is crucial in mobile health settings for accurate and reliable eyelid measurements. Deploying AI on mobile platforms significantly extends the accessibility and efficiency of oculoplastic diagnostics, meeting the increasing demand for advanced, readily available mobile health services.

| Backbone | DINO v2 | MRD1 | | | MRD2 | | | LF | | |
|---|---|---|---|---|---|---|---|---|---|---|
| | | MSE | MAE | $R^2$ | MSE | MAE | $R^2$ | MSE | MAE | $R^2$ |
| ViT-S | ✗ | 0.7414 | 0.7065 | 0.5206 | 0.6140 | 0.6220 | -0.0270 | 4.2521 | 1.5567 | -0.0323 |
| | ✓ | 0.6484 | 0.6446 | 0.5018 | 0.6658 | 0.6584 | -0.2499 | 3.8568 | 1.5795 | 0.2607 |
| ViT-B | ✗ | 0.7052 | 0.6554 | 0.5321 | 0.5858 | 0.5817 | 0.0954 | 6.4206 | 1.9121 | -0.4834 |
| | ✓ | 0.6087 | 0.5976 | 0.5208 | 0.4583 | 0.5514 | 0.0821 | 3.4134 | 1.4711 | 0.2676 |
| ViT-L | ✗ | 0.5974 | 0.6066 | 0.5416 | 0.5827 | 0.6030 | -0.0183 | 5.6000 | 1.7657 | -0.1210 |
| | ✓ | **0.5472** | **0.5957** | **0.6093** | **0.3769** | **0.4805** | **0.3639** | **3.3477** | **1.4327** | **0.3582** |

Table 4. Comparative Analysis of Self-Supervised Pre-training on Eyelid Measurement Predictions

**Scaling Efficiency and Stability in Model Performance**

Figure 12 presents model performance in terms of negative MSE across three backbone architectures—SE-ResNet, EfficientNet, and DINOv2—evaluated on MRD1, MRD2, and LF. While DINOv2 does not achieve the highest task-specific scores across all conditions, it demonstrates one of its most important strengths: scaling stability.

Consistent with findings from the original DINOv2 and Vision Transformer literature, our results confirm that as model complexity increases, particularly within the ViT family, performance becomes more stable across tasks. DINOv2 exhibits remarkably low performance variance between MRD1, MRD2, and LF, indicating that its self-supervised representations generalize effectively across different measurement types. This stands in contrast to SE-ResNet and EfficientNet, which, although competitive in specific tasks, show greater fluctuation in accuracy —especially in LF prediction, where modeling dynamic gaze-dependent anatomy is more complex.

The DINOv2 paper emphasizes the role of scaling model size and data volume to improve generalization and training stability. Our empirical results align with this: even without extensive fine-tuning, DINOv2 maintains consistent, high-quality performance across tasks, particularly when paired with lightweight regressors like MLP or Deep Ensemble. This predictable scaling behavior, coupled with its robustness in frozen or low-resource settings, makes DINOv2 an ideal candidate for deployment in clinical and mobile environments, where uniform reliability across multiple outputs is often more critical than peak performance in isolated tasks.

In summary, while SE-ResNet and EfficientNet show strength in specific scenarios, DINOv2's scaling efficiency and cross-task stability reaffirm the advantages of Vision Transformers for medical imaging. These properties are not only theoretical but practically observable in our results, supporting DINOv2 as a scalable and dependable foundation for automated eyelid measurement systems.

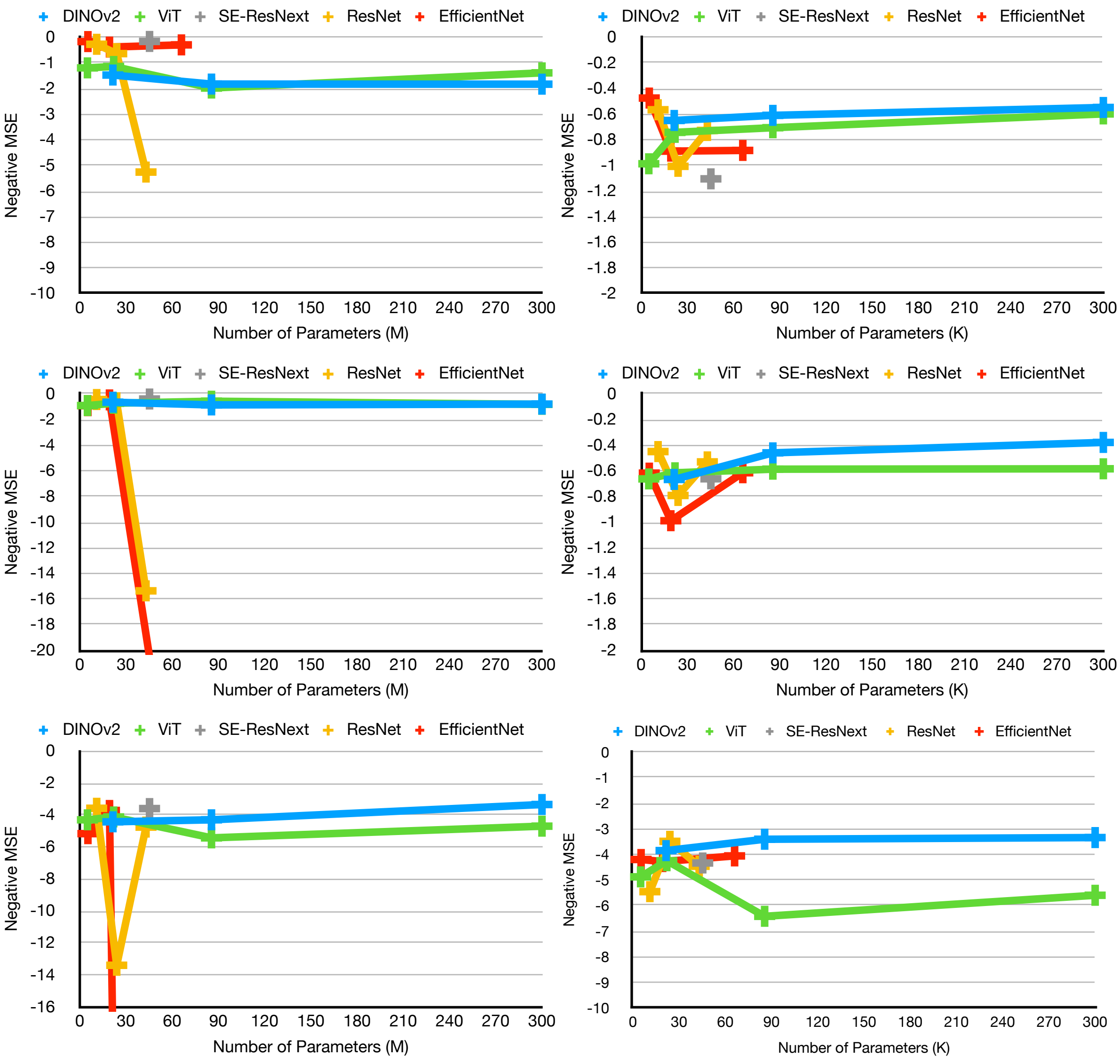

Figure 12. Scaling Properties for MRD1 (Top), MRD2 (Middle), LF (Bottom) Prediction: effect of model size on 1) end-to-end fine-tuning and 2) frozen.

## A Closer Look into Regressor

Figure 13 presents a comparative evaluation of multiple regression heads—MLP, iTransformer, FTTransformer, TabTransformer, and Deep Ensemble—across the three critical measurement tasks: MRD1, MRD2, and LF. Evaluations were conducted using three backbone networks (SE-ResNet, EfficientNet, and DINOv2), with Mean Squared Error (MSE) serving as the primary performance metric. In clinical regression tasks, lower MSE indicates higher precision, and consistency across tasks reflects a model's stability, which is crucial for real-world deployment.

As shown in Figure 13, the Deep Ensemble and MLP regressors emerged as the top performers across the majority of tasks, consistently achieving the lowest MSE scores. This demonstrates their superior capability in delivering precise, low-variance predictions, especially in the context of periocular measurements where anatomical variability and image noise can challenge model robustness. Deep Ensemble exhibited exceptional consistency and generalization, particularly when paired with DINOv2. Its averaged predictions not only reduced variance but also smoothed out noise in edge cases and underrepresented clinical examples. This led to stable, high-fidelity performance across MRD1, MRD2, and LF, making it the most reliable regressor overall. MLP, while simpler in design and more lightweight in computation, showed surprisingly competitive performance, especially in MRD1 and MRD2 prediction. Its minimalistic structure appears well-matched to the dense visual embeddings extracted by DINOv2, suggesting that when powered by high-quality features, even basic regressors can achieve state-of-the-art performance.

Beyond raw accuracy, inter-task stability is critical in clinical AI systems. Fluctuations in prediction quality across different tasks (e.g., MRD1 vs. LF) can erode clinical trust and limit deployment. According to Figure 13, Deep Ensemble maintained remarkably low variance in MSE across all three tasks and across all backbones—highlighting its value not only in accuracy but also in delivering consistent clinical performance. MLP also demonstrated strong task-level stability, particularly when combined with EfficientNet or DINOv2. Although it occasionally lagged behind in LF prediction—likely due to the added complexity of integrating up- and down-gaze information—it still outperformed more complex regressors like FTTransformer in terms of overall error consistency. In contrast, transformer-based regressors such as iTransformer, FTTransformer, and TabTransformer showed more variable task-level performance. While they performed competitively in certain configurations, their higher MSE and less stable behavior across tasks make them less favorable in contexts where uniform precision is essential.
From a clinical standpoint, these findings reinforce the importance of model simplicity, reliability, and stability over raw architectural complexity. Both MLP and Deep Ensemble offer a pragmatic and effective solution for mobile and clinical deployment—delivering high accuracy across a range of measurement tasks with minimal tuning or retraining.

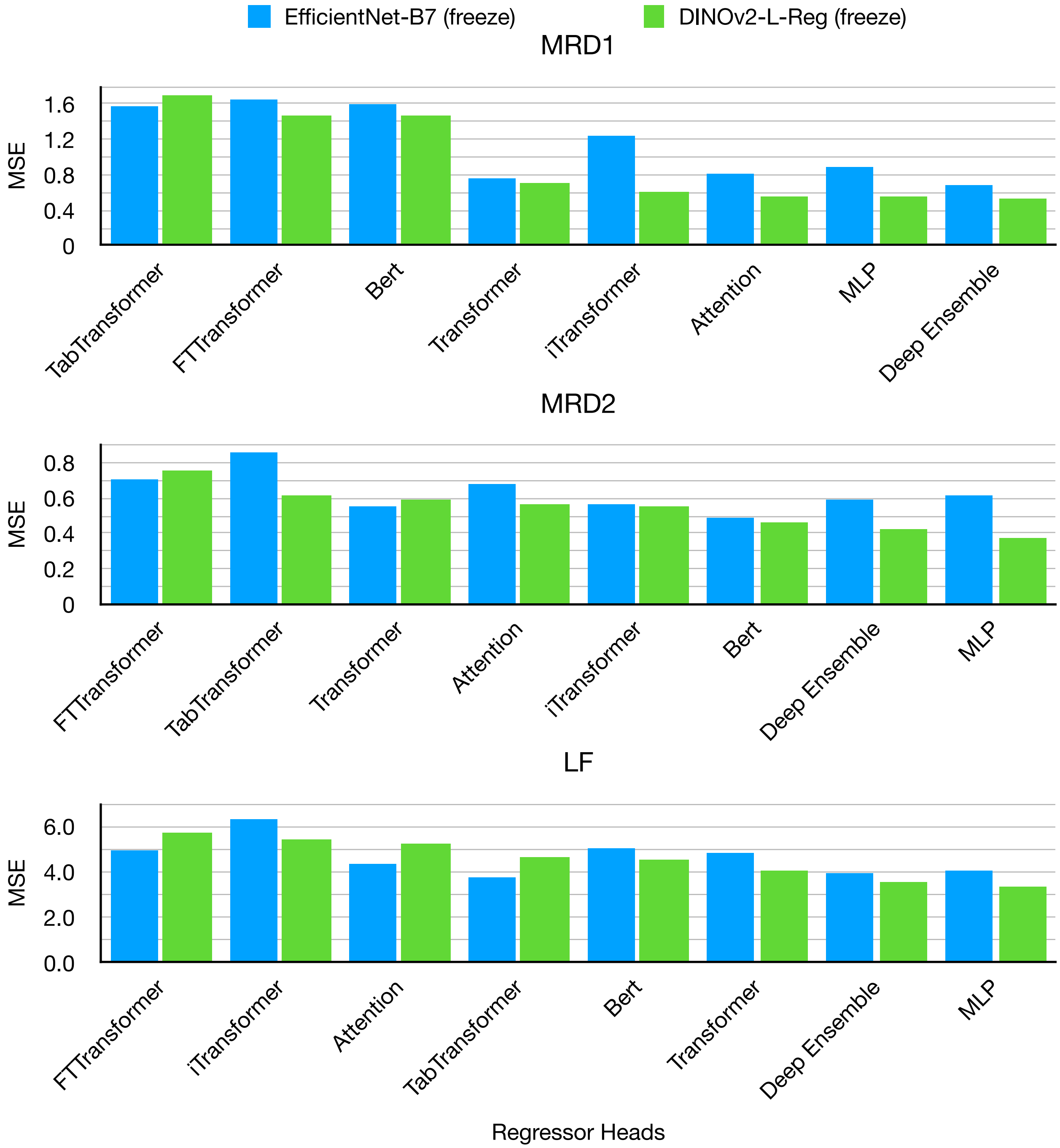

Figure 13. Comparative analysis of regression heads (MLP, iTransformer, FTTransformer, TabTransformer, Deep Ensemble) applied to MRD1, MRD2, and LF prediction tasks using three different backbones (SE-ResNet, EfficientNet, DINOv2). MSE values are reported for each configuration. Deep Ensemble and MLP regressors consistently yield the lowest MSE and highest inter-task stability, establishing them as the most clinically viable choices for eyelid measurement systems.

**Validation on Focal Loss, Orthogonality, and Binary Encoding**

Figure 14 evaluates the effectiveness of three targeted learning strategies—focal loss, orthogonality regularization (OR), and binary encoding—in improving model stability and regression performance. Each method was designed to address a specific challenge: focal loss for data imbalance, OR for multi-task interference, and binary encoding for regression precision and training robustness.

Focal loss, while conceptually effective at addressing long-tailed clinical distributions, showed unstable behavior when applied in isolation. In some trials, especially with LF prediction, training became erratic, with increased variance in loss curves. However, when focal loss was combined with orthogonality regularization, model stability improved significantly. OR constrained gradient directions across MRD1, MRD2, and LF, reducing interference between tasks and helping smooth the sharp optimization dynamics introduced by focal loss. This pairing yielded more reliable convergence, especially in transformer-based models like DINOv2.

Orthogonality regularization alone also contributed meaningfully to performance consistency. By decoupling task-specific updates in the parameter space, it reduced overfitting to dominant outputs and improved generalization across all three tasks. This effect was particularly notable in DINOv2 and EfficientNet, where cross-task stability is crucial for clinical reliability.

The most striking effect came from binary encoding, which converted regression targets into multi-bit classification tasks. This reformulation helped stabilize training, particularly in EfficientNet, where standard regression loss led to loss explosions in MRD2 and LF. Binary encoding provided smoother gradients and simplified the optimization landscape, enabling the backbone to learn more effectively. It also allowed controllable prediction precision, aligning well with clinical needs for fine-grained but stable output.

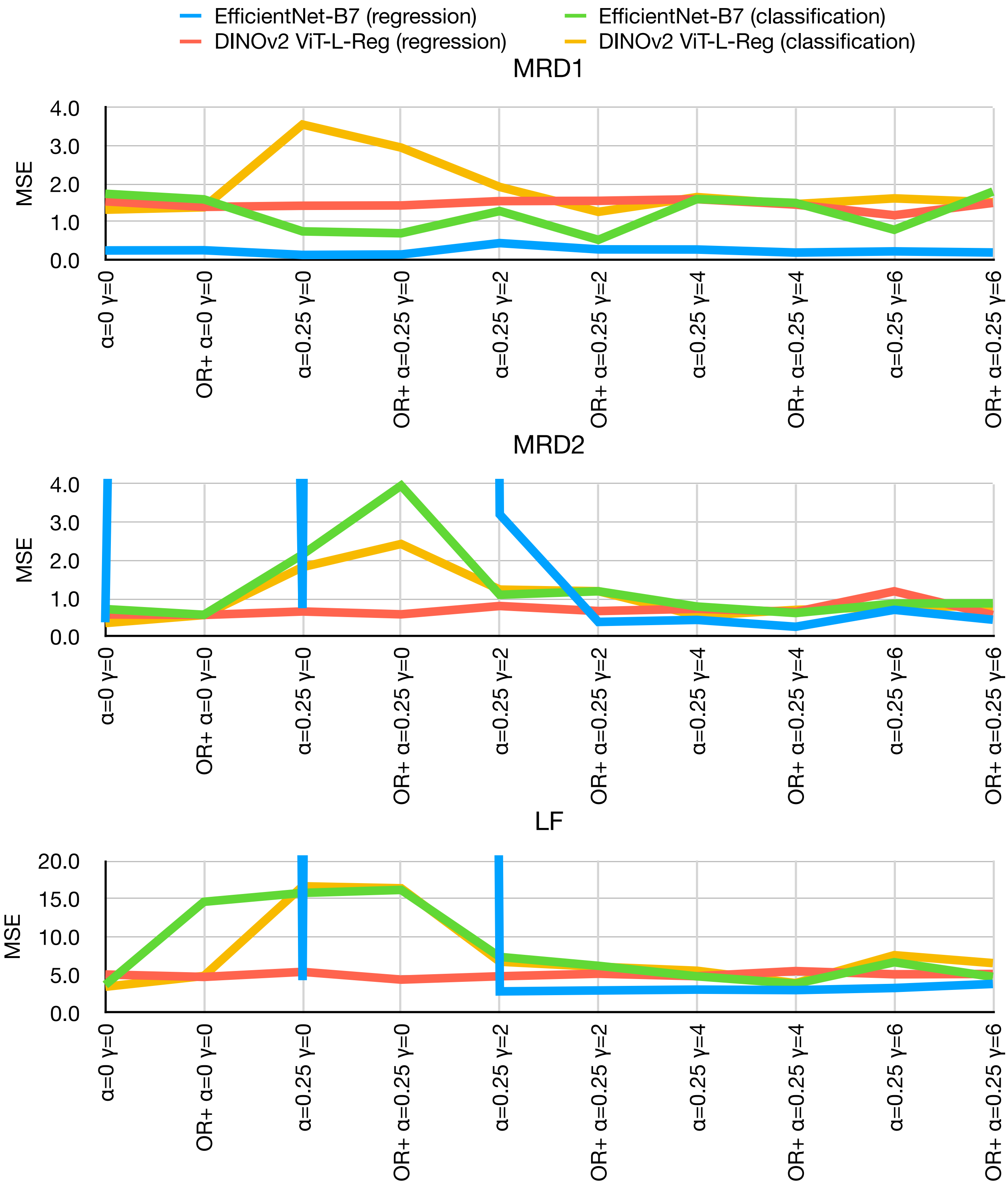

Figure 14. Effect of focal loss, orthogonality regularization (OR), and binary encoding on negative MSE across MRD1, MRD2, and LF. While focal loss alone introduced some instability, combining it with OR improved consistency. Binary encoding notably stabilized training, especially in EfficientNet, preventing loss spikes in MRD2 and LF.

**Efforts on Domain-Specific Self-Supervision and Feature Pyramid Architecture**

Figure 15 presents the impact of applying domain-specific self-supervised learning and feature pyramid architecture (FPN) to enhance model performance on MRD1, MRD2, and LF. These efforts were motivated by the need to adapt general vision models to the fine-scale anatomical structures and task-specific constraints of eyelid measurement.

Due to GPU memory limitations, domain-specific self-supervised training was not implemented for larger models such as EfficientNet-B4 and DINOv2-B, limiting our evaluation to pre-existing pretrained representations. For DINOv2 in particular, which already benefits from high-quality self-supervised learning at scale, we observed no consistent performance gain from continued domain-specific pretraining. This may be attributed to the relatively small and low-diversity corpus of ophthalmic images used for adaptation, which may not have offered enough variation to meaningfully improve feature representation. As a result, the added self-supervision did not significantly boost accuracy across tasks, highlighting the importance of data scale and diversity in successful SSL adaptation.

In contrast, the incorporation of Feature Pyramid Networks showed more promising results. While FPN did not consistently improve performance across all backbones or all tasks, its integration with DINOv2 led to robust and repeatable gains, particularly in MRD1 and MRD2. These tasks require precise localization of eyelid margins and tracking subtle positional changes between gaze directions—both of which benefit from multi-scale spatial encoding. In contrast to the results from [ViT PFN], FPN's ability to merge high-level vision transformers' semantic features with low-level spatial detail likely contributed to this improvement.

Overall, Figure 15 illustrates that while domain-specific self-supervision showed limited success under current constraints, FPN consistently enhanced spatial awareness when paired with transformer-based models like DINOv2. These findings suggest that architectural enhancements like FPN may offer more immediate and scalable benefits for clinical deployment, especially in scenarios where annotated data is scarce and model memory is constrained.

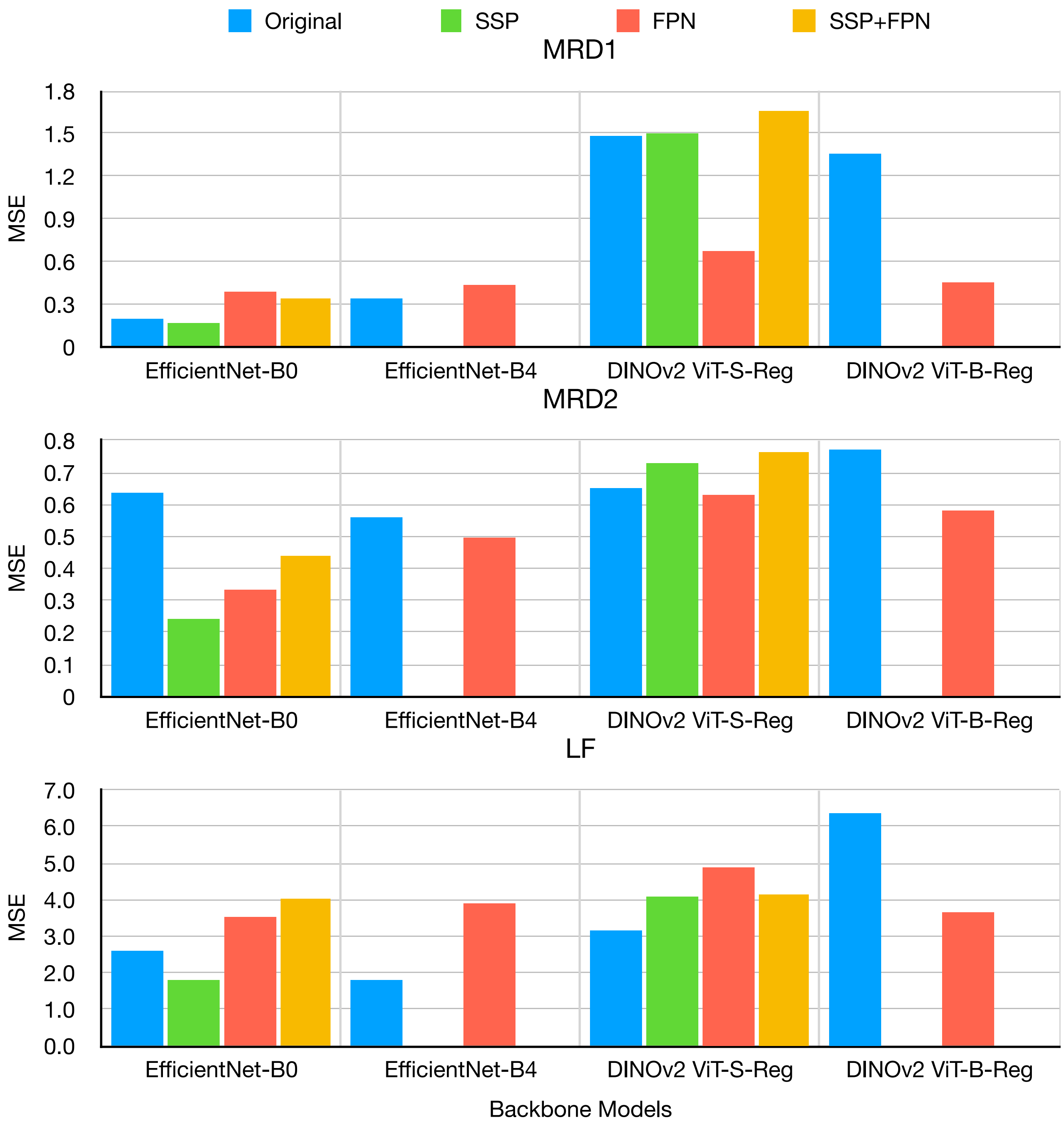

Figure 15. Comparison of model performance with and without domain-specific self-supervised learning and feature pyramid architecture (FPN) across MRD1, MRD2, and LF tasks. Due to GPU memory limits, self-supervised training was not applied to EfficientNet-B4 and DINOv2-B. Overall, self-supervision yielded limited benefit, likely due to the low diversity and scale of the ophthalmic image corpus. In contrast, FPN provided consistent improvements when integrated with DINOv2, particularly in MRD2 and LF, highlighting its effectiveness in enhancing spatial feature representation in transformer-based models.

# Conclusion

This study presents a comprehensive framework for automated eyelid measurement using deep learning, emphasizing performance, efficiency, and adaptability for mobile health applications. By comparing SE-ResNet, EfficientNet, and DINOv2 architectures, we demonstrate that DINOv2, particularly when paired with lightweight regression heads like MLP and Deep Ensemble, consistently delivers high-precision predictions for MRD1, MRD2, and LF. Its robustness under frozen configurations also highlights its suitability for mobile deployment.

To address the clinical challenges of data imbalance and multi-task learning, we introduced three targeted learning strategies. Focal loss, though promising for emphasizing minority cases, proved unstable in isolation. However, its combination with orthogonality regularization significantly enhanced training stability by reducing task interference. Binary encoding emerged as the most impactful strategy—transforming regression into fine-grained classification tasks, smoothing optimization, and notably improving performance, especially in EfficientNet.

Furthermore, domain-specific self-supervised learning provided limited benefits under current constraints, primarily due to the scale and diversity of the unlabeled ophthalmic image corpus. In contrast, Feature Pyramid Networks (FPN) proved to be a consistently effective architectural enhancement. When integrated with DINOv2, FPN improved spatial awareness and anatomical localization, yielding measurable gains in MRD1 and MRD2 prediction.

In summary, this work validates the strategic synergy of vision transformers, architectural enhancements, and task-specific learning strategies in enabling accurate, scalable, and mobile-ready eyelid measurement systems. These findings not only support DINOv2's clinical viability but also highlight pathways for future research in foundation model adaptation, lightweight deployment, and reliable AI integration in ophthalmology and beyond.

## Supplementary Data

| Model | #Param | MRD1 | | | MRD2 | | | LF | | |
|---|---|---|---|---|---|---|---|---|---|---|
| | | MSE | MAE | $R^2$ | MSE | MAE | $R^2$ | MSE | MAE | $R^2$ |
| EfficientNet B0 | 5M | 0.1637 | 0.3081 | 0.8863 | 0.9033 | 0.8144 | -0.3982 | 5.1682 | 1.9680 | -0.1312 |
| | 274K | 0.4737 | 0.5360 | 0.6911 | 0.6166 | 0.6165 | 0.0003 | 4.2038 | 1.6468 | 0.2060 |
| EfficientNet B4 | 19M | 0.3789 | 0.4839 | 0.7628 | 0.4508 | 0.5202 | 0.3523 | 3.7202 | 1.5796 | 0.0071 |
| | 274K | 0.8880 | 0.7688 | 0.4026 | 0.9880 | 0.7783 | -0.4681 | 4.2842 | 1.5593 | 0.0676 |
| EfficientNet B7 | 66M | 0.2925 | 0.4196 | 0.7974 | 36.4946 | 2.4391 | -54.6941 | 312.77 | 6.6107 | -70.368 |
| | 274K | 0.8825 | 0.7377 | 0.4445 | 0.6136 | 0.5973 | 0.1794 | 4.0636 | 1.5845 | 0.1026 |
| ResNet18 | 11M | 0.2695 | 0.4155 | 0.7894 | 0.4205 | 0.4998 | 0.3891 | 3.5831 | 1.5323 | 0.0938 |
| | 274K | 0.5652 | 0.6061 | 0.6775 | 0.4483 | 0.5496 | 0.0779 | 5.4584 | 1.8946 | -0.2393 |
| ResNet50 | 24M | 0.6338 | 0.6264 | 0.5431 | 0.6779 | 0.6467 | -0.3156 | 13.3881 | 2.3257 | -2.3682 |
| | 274K | 1.0064 | 0.8003 | 0.3780 | 0.7908 | 0.7218 | -0.3706 | 3.4956 | 1.4896 | 0.1506 |
| ResNet101 | 43M | 5.2572 | 1.1557 | -2.7312 | 15.3555 | 2.0428 | -18.4985 | 4.7583 | 1.7815 | -0.1180 |
| | 274K | 0.7266 | 0.6631 | 0.3623 | 0.5290 | 0.5489 | 0.2237 | 4.4669 | 1.6466 | 0.0508 |
| SE-ResNext101 | 45M | **0.1583** | 0.2996 | 0.9046 | 0.3805 | 0.4482 | 0.4615 | 3.6013 | 1.5114 | 0.0837 |
| | 274K | 1.1043 | 0.8297 | 0.2942 | 0.6608 | 0.6453 | -0.1207 | 4.3397 | 1.6831 | -0.0430 |

Table 5-1. Assessment of Convolution-based Models for MRD1, MRD2, and LF tasks.

| Model | #Param | MRD1 | | | MRD2 | | | LF | | |
|---|---|---|---|---|---|---|---|---|---|---|
| | | MSE | MAE | $R^2$ | MSE | MAE | $R^2$ | MSE | MAE | $R^2$ |
| ViT-T | 5M | 1.1903 | 0.8683 | 0.1640 | 0.8902 | 0.7893 | -0.6504 | 4.3038 | 1.5952 | -0.0220 |
| | 274K | 0.9866 | 0.8285 | 0.2992 | 0.6616 | 0.6450 | 0.1467 | 4.8808 | 1.6461 | -0.1486 |
| ViT-S | 22M | 1.1358 | 0.8422 | 0.0739 | 0.7099 | 0.6128 | -0.0870 | 4.1253 | 1.5459 | -0.0019 |
| | 274K | 0.7414 | 0.7065 | 0.5206 | 0.6140 | 0.6220 | -0.0270 | 4.2521 | 1.5567 | -0.0323 |
| ViT-B | 86M | 1.9770 | 1.1496 | -0.1093 | 0.5514 | 0.5851 | -0.0081 | 5.4303 | 1.7205 | -0.0693 |
| | 274K | 0.7052 | 0.6554 | 0.5321 | 0.5858 | 0.5817 | 0.0954 | 6.4206 | 1.9121 | -0.4834 |
| ViT-L | 300M | 1.3846 | 0.9382 | 0.0000 | 0.7960 | 0.7313 | -0.2183 | 4.7010 | 1.6840 | -0.1452 |
| | 274K | 0.5974 | 0.6066 | 0.5416 | 0.5827 | 0.6030 | -0.0183 | 5.6000 | 1.7657 | -0.1210 |
| DINOv2 ViT-S-Reg | 21M | 1.4700 | 0.9859 | -0.0088 | 0.6292 | 0.6211 | -0.0024 | 4.4282 | 1.6696 | -0.3186 |
| | 116K | 0.6484 | 0.6446 | 0.5018 | 0.6658 | 0.6584 | -0.2499 | 3.8568 | 1.5795 | 0.2607 |
| DINOv2 ViT-B-Reg | 86M | 1.8224 | 1.1345 | -0.0283 | 0.8410 | 0.7335 | -0.2695 | 4.3026 | 1.6934 | -0.2419 |
| | 214K | 0.6087 | 0.5976 | 0.5208 | 0.4583 | 0.5514 | 0.0821 | 3.4134 | 1.4711 | 0.2676 |
| DINOv2 ViT-L-Reg | 300M | 1.8270 | 1.0974 | -0.2000 | 0.7767 | 0.6590 | -0.0001 | 3.3484 | 1.4205 | -0.0271 |
| | 280K | 0.5472 | 0.5957 | 0.6093 | **0.3769** | 0.4805 | 0.3639 | **3.3477** | 1.4327 | 0.3582 |

Table 5-2. Assessment of Transformer-based Models for MRD1, MRD2, and LF tasks.

| Model | Freeze | Head | MRD1 | | | MRD2 | | | LF | | |
|---|---|---|---|---|---|---|---|---|---|---|---|
| | | | MSE | MAE | $R^2$ | MSE | MAE | $R^2$ | MSE | MAE | $R^2$ |
| ResNet18 | ✗ | BNN | 167.0166 | 11.3116 | -108.8013 | 83.4387 | 8.1088 | -120.2071 | 44.5743 | 3.7291 | -8.7133 |
| | | Attention | 0.203 | 0.3286 | 0.8511 | 0.3086 | 0.439 | 0.5005 | 2.6547 | 1.2523 | 0.4184 |
| | | Transformer | 0.1695 | 0.3143 | 0.8823 | 0.2017 | 0.3561 | 0.684 | 2.0687 | 1.1082 | 0.4828 |
| | | iTransformer | **0.134** | 0.2858 | 0.8849 | 0.349 | 0.4157 | 0.5696 | 2.5334 | 1.2448 | 0.2953 |
| | | Bert | 0.1816 | 0.3373 | 0.8769 | 0.5651 | 0.5919 | -0.0275 | 5.1629 | 1.745 | -0.0777 |
| | | FTTransformer | 1.7739 | 1.1125 | -0.0047 | 0.6474 | 0.6258 | -0.0078 | 4.8384 | 1.7483 | -0.1303 |
| | | TabTransformer | 1.4761 | 0.9749 | -0.0434 | 0.5896 | 0.5878 | -0.0421 | 3.8674 | 1.5921 | -0.0148 |
| | | DeepEnsemble | 0.2939 | 0.4112 | 0.824 | 0.2502 | 0.387 | 0.5727 | 2.518 | 1.2438 | 0.4465 |
| | ✓ | BNN | 8086.4424 | 79.3351 | -4960.9246 | 2776.2603 | 41.5799 | -5036.8714 | 1217.4401 | 28.1664 | -249.7187 |
| | | Attention | 0.7583 | 0.6568 | 0.4495 | 0.7445 | 0.6529 | 0.0673 | 3.5019 | 1.4894 | 0.1373 |
| | | Transformer | 0.8073 | 0.7031 | 0.3972 | 0.4624 | 0.534 | 0.2176 | 3.8378 | 1.6146 | 0.0331 |
| | | iTransformer | 0.7139 | 0.6535 | 0.5459 | 0.7553 | 0.6566 | -0.1425 | 3.8082 | 1.5471 | 0.0657 |
| | | Bert | 1.4141 | 0.9693 | -0.0021 | 0.8072 | 0.6612 | -0.0036 | 5.4357 | 1.775 | -0.1551 |
| | | FTTransformer | 1.5237 | 0.9998 | 0.0003 | 0.7895 | 0.6697 | -0.0037 | 5.2923 | 1.8709 | -0.0916 |
| | | TabTransformer | 1.5078 | 1.0119 | -0.01 | 0.5774 | 0.5859 | -0.0047 | 5.0127 | 1.7866 | -0.0069 |
| | | DeepEnsemble | 0.6727 | 0.6326 | 0.5488 | 0.5461 | 0.5554 | 0.2314 | 4.4414 | 1.6085 | 0.2267 |
| ResNet50 | ✗ | BNN | 6.9715 | 2.2983 | -3.9176 | 119.0243 | 7.2884 | -169.6483 | 5.7059 | 1.972 | -0.345 |
| | | Attention | 0.2436 | 0.3714 | 0.8173 | 0.2774 | 0.3905 | 0.6122 | 2.1214 | 1.1433 | 0.5171 |
| | | Transformer | 0.2065 | 0.3246 | 0.8663 | 0.2318 | 0.3735 | 0.6483 | 3.4179 | 1.5149 | 0.3154 |
| | | iTransformer | 0.2297 | 0.3431 | 0.8454 | 0.2306 | 0.3801 | 0.6868 | 2.236 | 1.1369 | 0.506 |
| | | Bert | 1.4846 | 1.0193 | -0.0009 | 0.4386 | 0.5444 | 0.2807 | 5.7846 | 1.8178 | -0.0932 |
| | | FTTransformer | 1.245 | 0.9144 | -0.0025 | 0.6387 | 0.5945 | -0.0264 | 5.2112 | 1.8143 | -0.1823 |
| | | TabTransformer | 1.3723 | 0.9345 | -0.0182 | 0.7595 | 0.6594 | -0.0515 | 3.8493 | 1.5578 | -0.0516 |
| | | DeepEnsemble | 0.3465 | 0.4604 | 0.7256 | 0.2111 | 0.3457 | 0.5888 | 1.8269 | 1.0595 | 0.6271 |
| | | BNN | 8111.708 | 78.8948 | -5777.6203 | 4157.3511 | 57.0742 | -6049.9984 | 1234.7822 | 28.5265 | -249.1676 |
| | | Attention | 0.6512 | 0.6582 | 0.5199 | 0.501 | 0.5677 | -0.0198 | 5.1663 | 1.8402 | -0.1232 |
| | ✓ | Transformer | 0.9171 | 0.7037 | 0.3891 | 0.7303 | 0.6405 | 0.2656 | 4.9957 | 1.7335 | -0.0118 |
| | | iTransformer | 1.5008 | 0.9744 | -0.009 | 0.541 | 0.5988 | 0.0787 | 5.8998 | 1.8024 | -0.1414 |
| | | Bert | 1.6615 | 1.0569 | -0.0075 | 0.6987 | 0.6781 | -0.1572 | 5.1683 | 1.6341 | -0.0755 |
| | | FTTransformer | 1.5752 | 1.029 | -0.0008 | 0.5009 | 0.5659 | -0.0101 | 5.8295 | 1.8949 | -0.1882 |
| | | TabTransformer | 1.2836 | 0.9466 | 0.0048 | 0.5831 | 0.5855 | -0.0365 | 4.7704 | 1.6783 | 0.0131 |
| | | DeepEnsemble | 0.774 | 0.7111 | 0.537 | 0.5605 | 0.5916 | 0.0723 | 5.4886 | 1.8848 | -0.0234 |
| ResNet101 | ✗ | BNN | 133.1467 | 3.531 | -92.5011 | 22.9168 | 4.3519 | -40.2289 | 6.2921 | 1.8991 | -0.5268 |
| | | Attention | 0.2826 | 0.3776 | 0.8174 | 0.2374 | 0.3652 | 0.6271 | 1.7417 | 0.9839 | 0.5266 |
| | | Transformer | 0.191 | 0.2963 | 0.8432 | 0.2847 | 0.4427 | 0.4272 | 2.3316 | 1.1532 | 0.3834 |
| | | iTransformer | 0.4028 | 0.5191 | 0.7362 | 0.9083 | 0.7523 | -0.4103 | 6.0823 | 1.8159 | -0.2116 |
| | | Bert | 1.6521 | 1.0834 | -0.1272 | 0.7486 | 0.6389 | -0.0371 | 4.3975 | 1.625 | -0.1511 |
| | | FTTransformer | 1.4625 | 0.9894 | -0.0044 | 0.6938 | 0.6105 | -0.0036 | 5.6534 | 1.8935 | -0.3145 |
| | | TabTransformer | 1.4984 | 1.0003 | -0.0031 | 0.7223 | 0.6327 | -0.0825 | 4.0285 | 1.5962 | -0.0232 |
| | | DeepEnsemble | 0.1827 | 0.3136 | 0.8633 | **0.1462** | 0.2994 | 0.75 | 1.7209 | 1.0516 | 0.5313 |
| | ✓ | BNN | 17136.3516 | 121.003 | 12206.181 | 0327.3652 | 95.3179 | 19373.706 | 1569.3082 | 30.7897 | -398.4597 |
| | | Attention | 0.736 | 0.6977 | 0.4754 | 0.3538 | 0.4698 | 0.1218 | 5.5996 | 1.8635 | -0.0128 |
| | | Transformer | 1.1168 | 0.8368 | 0.1423 | 0.7276 | 0.6243 | 0.0104 | 4.5076 | 1.7164 | -0.0403 |
| | | iTransformer | 1.5674 | 1.0323 | -0.0059 | 0.6183 | 0.6214 | -0.2038 | 4.347 | 1.632 | -0.0186 |
| | | Bert | 1.4755 | 1.0112 | 0.0001 | 0.98 | 0.7372 | -0.0125 | 4.2329 | 1.6026 | -0.1062 |
| | | FTTransformer | 1.7031 | 1.0732 | -0.0004 | 0.4694 | 0.5401 | -0.0012 | 4.8967 | 1.7096 | -0.0506 |
| | | TabTransformer | 1.5073 | 0.9964 | -0.0986 | 0.6064 | 0.5946 | -0.0803 | 5.5792 | 1.7875 | 0.0107 |
| | | DeepEnsemble | 0.7649 | 0.6937 | 0.4666 | 0.5486 | 0.58 | 0.0604 | 4.9905 | 1.6989 | 0.064 |
| | ✗ | BNN | 43.4415 | 5.4586 | -31.1209 | 39.5175 | 4.832 | -69.9858 | 73.5463 | 6.7826 | -14.381 |
| | | Attention | 0.2128 | 0.3618 | 0.8526 | 0.2223 | 0.3702 | 0.6454 | **1.6829** | 1.0498 | 0.6027 |
| | | Transformer | 0.2516 | 0.362 | 0.8411 | 0.2484 | 0.3872 | 0.6554 | 2.3508 | 1.223 | 0.446 |
| | | iTransformer | 0.3358 | 0.4219 | 0.7946 | 0.2494 | 0.3955 | 0.5557 | 5.2548 | 1.7355 | -0.1117 |
| | | Bert | 0.2122 | 0.3354 | 0.8483 | 0.4294 | 0.4989 | 0.4456 | 4.9072 | 1.7613 | -0.0911 |
| | | FTTransformer | 1.641 | 1.0491 | -0.0416 | 0.6791 | 0.6273 | -0.0069 | 5.6809 | 1.9631 | -0.2091 |
| EfficientNet B0 | | TabTransformer | 1.3044 | 0.9406 | -0.0211 | 0.6428 | 0.621 | -0.0295 | 4.0465 | 1.5791 | -0.0358 |
| | | DeepEnsemble | 0.3011 | 0.3972 | 0.7774 | 0.2693 | 0.4226 | 0.5381 | 2.1036 | 1.1293 | 0.4782 |
| | ✓ | BNN | 1064.5153 | 26.1478 | -879.9711 | 7343.168 | 79.2973 | 12766.419 | 378.1581 | 15.3522 | -75.0563 |
| | | Attention | 0.4324 | 0.4777 | 0.7 | 0.438 | 0.5289 | 0.205 | 3.9611 | 1.5852 | 0.1525 |
| | | Transformer | 0.6041 | 0.5896 | 0.585 | 0.4733 | 0.4854 | 0.4055 | 4.0514 | 1.5736 | -0.0006 |
| | | iTransformer | 0.5884 | 0.5772 | 0.5675 | 0.5513 | 0.6054 | 0.1094 | 4.4539 | 1.6654 | -0.0415 |
| | | Bert | 1.4339 | 0.9829 | -0.0021 | 0.5437 | 0.5618 | 0.2946 | 4.4757 | 1.6508 | -0.0038 |
| | | FTTransformer | 1.5194 | 1.0177 | 0.0011 | 0.5792 | 0.5584 | -0.0047 | 5.2181 | 1.8236 | -0.1893 |
| | | TabTransformer | 1.2533 | 0.9265 | -0.0087 | 0.7478 | 0.659 | -0.0023 | 5.1923 | 1.7416 | -0.0103 |
| | | DeepEnsemble | 0.4591 | 0.5442 | 0.6755 | 0.5153 | 0.526 | 0.284 | 3.586 | 1.5343 | 0.1147 |
| EfficientNet B4 | ✗ | BNN | 11.0866 | 2.6787 | -6.8907 | 17.7613 | 3.3456 | -23.9943 | 14.7434 | 3.1114 | -2.2936 |
| | | Attention | 0.2926 | 0.4079 | 0.8255 | 0.3974 | 0.4641 | 0.4887 | 6.2035 | 2.084 | -0.215 |
| | | Transformer | 0.274 | 0.3816 | 0.8226 | 0.3169 | 0.4102 | 0.5523 | 3.3557 | 1.4351 | 0.3895 |
| | | iTransformer | 0.2664 | 0.3906 | 0.8283 | 0.2304 | 0.3792 | 0.569 | 4.0332 | 1.5817 | 0.0845 |
| | | Bert | 0.363 | 0.4893 | 0.7565 | 0.3512 | 0.4613 | 0.4989 | 4.906 | 1.6534 | -0.0806 |
| | | TabTransformer | 1.8247 | 1.096 | -0.0449 | 0.6289 | 0.6417 | -0.0802 | 6.9275 | 2.099 | -0.0451 |
| | | DeepEnsemble | 0.2749 | 0.3851 | 0.8172 | 0.3521 | 0.428 | 0.5392 | 2.9545 | 1.3154 | 0.3436 |
| | ✓ | BNN | 803.2334 | 20.9807 | -455.7572 | 601.1345 | 21.5813 | -930.0329 | 1739.0376 | 39.0472 | -408.4559 |
| | | Attention | 0.9306 | 0.7313 | 0.4369 | 0.8572 | 0.6973 | 0.0975 | 4.4926 | 1.6873 | -0.1253 |
| | | Transformer | 0.9639 | 0.7939 | 0.3737 | 0.6528 | 0.5934 | 0.1854 | 4.8517 | 1.7531 | 0.0022 |
| | | iTransformer | 1.5543 | 1.0025 | -0.162 | 0.5709 | 0.5644 | 0.0711 | 5.0262 | 1.8223 | -0.086 |
| | | Bert | 1.1072 | 0.8426 | 0.0043 | 0.5488 | 0.5852 | -0.0052 | 5.0418 | 1.5986 | -0.1064 |
| | | FTTransformer | 1.3956 | 0.9697 | -0.0177 | 0.6836 | 0.6353 | -0.0042 | 6.2059 | 1.9753 | -0.3229 |
| | | TabTransformer | 1.5187 | 1.0272 | -0.0091 | 0.6533 | 0.6254 | -0.0072 | 4.9261 | 1.7052 | -0.0251 |
| | | DeepEnsemble | 0.7598 | 0.7036 | 0.5212 | 0.8171 | 0.6604 | 0.0839 | 4.5471 | 1.6824 | -0.0953 |
| | | BNN | 2573.804 | 47.6175 | -1897.287 | 445.7147 | 15.9793 | -730.8379 | 552.6542 | 18.7015 | -115.5141 |
| | | Attention | 0.8025 | 0.7243 | 0.4269 | 0.6748 | 0.632 | 0.1188 | 4.3829 | 1.6145 | 0.0296 |
| | | Transformer | 0.7507 | 0.6811 | 0.4349 | 0.5465 | 0.5764 | -0.0168 | 4.8061 | 1.6649 | 0.0029 |
| EfficientNet B7 | ✓ | iTransformer | 1.246 | 0.9012 | 0.1483 | 0.5678 | 0.5986 | 0.0765 | 6.3286 | 1.9301 | -0.0271 |
| | | Bert | 1.604 | 1.0705 | -0.0142 | 0.4942 | 0.5788 | -0.0062 | 5.0341 | 1.7653 | -0.0459 |
| | | FTTransformer | 1.6621 | 1.0782 | -0.0007 | 0.7028 | 0.6477 | -0.0166 | 4.9491 | 1.756 | -0.1744 |
| | | TabTransformer | 1.5707 | 0.9902 | -0.0207 | 0.8513 | 0.6533 | -0.0374 | 3.7765 | 1.5471 | 0.0023 |
| | | DeepEnsemble | 0.6789 | 0.6449 | 0.4204 | 0.588 | 0.5748 | -0.0506 | 3.921 | 1.5081 | 0.1645 |
| ViT-T | ✗ | BNN | 18.7959 | 4.1412 | -10.4386 | 8.0598 | 2.0012 | -10.7722 | 11.3355 | 2.7491 | -1.6023 |
| | | Attention | 0.3176 | 0.4068 | 0.7968 | 0.5558 | 0.5829 | 0.046 | 3.3508 | 1.4483 | 0.1469 |
| | | Transformer | 0.6064 | 0.6247 | 0.6013 | 0.7126 | 0.6318 | -0.018 | 4.3457 | 1.5975 | -0.0023 |
| | | iTransformer | 1.7772 | 1.0812 | -0.0335 | 0.6808 | 0.6238 | -0.1093 | 4.2339 | 1.6607 | -0.0301 |
| | | Bert | 1.5099 | 1.0146 | -0.0013 | 0.5729 | 0.5964 | -0.1635 | 4.0734 | 1.5669 | -0.0271 |
| | | FTTransformer | 1.503 | 0.9898 | -0.0065 | 0.6375 | 0.6309 | -0.0253 | 6.1337 | 2.0744 | -0.1299 |
| | | TabTransformer | 1.4567 | 0.9928 | -0.0029 | 0.6359 | 0.5984 | -0.0266 | 4.5666 | 1.6945 | -0.0051 |
| | | DeepEnsemble | 0.3774 | 0.4573 | 0.7312 | 0.3902 | 0.4426 | 0.2945 | 4.9561 | 1.7271 | -0.0005 |
| | ✓ | BNN | 1104.5004 | 25.2701 | -742.0118 | 845.4702 | 23.0112 | -1210.3936 | 1658.7695 | 33.7601 | -365.4707 |
| | | Attention | 1.0975 | 0.8202 | 0.2918 | 0.8188 | 0.6568 | -0.0658 | 4.1844 | 1.5872 | -0.0197 |
| | | Transformer | 0.9593 | 0.7449 | 0.3232 | 0.5056 | 0.5822 | -0.0779 | 5.1554 | 1.7916 | 0.0014 |
| | | iTransformer | 1.0709 | 0.8142 | 0.2802 | 0.6989 | 0.6569 | -0.0017 | 4.1069 | 1.6063 | -0.2401 |
| | | Bert | 1.458 | 0.9857 | 0 | 0.6703 | 0.6081 | -0.0435 | 5.1328 | 1.7923 | -0.0725 |
| | | FTTransformer | 1.488 | 1.0055 | -0.0268 | 0.9045 | 0.6867 | -0.001 | 4.4428 | 1.7011 | -0.2659 |
| | | TabTransformer | 1.3088 | 0.9185 | -0.0045 | 0.6908 | 0.6164 | -0.0283 | 3.8166 | 1.4679 | -0.0023 |
| | | DeepEnsemble | 0.9477 | 0.7873 | 0.309 | 0.6714 | 0.6139 | 0.0579 | 5.2651 | 1.7479 | -0.0308 |
| | ✗ | BNN | 102.8181 | 10.0651 | -80.5179 | 60.4094 | 7.5721 | -100.6099 | 46.6829 | 6.4913 | -9.1668 |
| | | Attention | 0.258 | 0.3707 | 0.8415 | 0.4087 | 0.4455 | 0.3221 | 4.0468 | 1.61 | -0.0392 |
| | | Transformer | 0.3374 | 0.4226 | 0.7697 | 0.6376 | 0.5894 | -0.0387 | 4.077 | 1.7087 | -0.0014 |
| | | iTransformer | 1.2803 | 0.9244 | -0.0006 | 0.786 | 0.6255 | -0.0104 | 5.1076 | 1.732 | -0.1829 |
| | | Bert | 1.708 | 1.0701 | -0.0078 | 0.7697 | 0.6394 | -0.0379 | 4.4188 | 1.6384 | -0.0296 |
| | | FTTransformer | 1.4894 | 0.982 | -0.0057 | 0.9789 | 0.6962 | 0 | 5.4822 | 1.9252 | -0.351 |
| | | TabTransformer | 1.2866 | 0.9064 | 0.0002 | 0.608 | 0.5995 | -0.0007 | 5.5959 | 1.8401 | -0.069 |
| ViT-S | | DeepEnsemble | 0.2889 | 0.3951 | 0.7997 | 0.3233 | 0.4384 | 0.4109 | 5.0844 | 1.8533 | -0.0547 |
| | ✓ | BNN | 1232.8457 | 28.9547 | -772.7844 | 1012.9025 | 26.3789 | -1262.8942 | 553.6093 | 19.5825 | -110.9304 |
| | | Attention | 0.6859 | 0.6702 | 0.5148 | 0.5071 | 0.5551 | 0.206 | 4.3436 | 1.574 | -0.0242 |
| | | Transformer | 0.8468 | 0.7144 | 0.4482 | 0.509 | 0.5715 | 0.2572 | 4.5699 | 1.6998 | 0.0027 |
| | | iTransformer | 0.8837 | 0.7622 | 0.365 | 0.8521 | 0.6841 | -0.2287 | 5.7803 | 1.8964 | -0.041 |
| | | Bert | 1.4036 | 0.983 | -0.0231 | 0.7021 | 0.606 | -0.0555 | 5.0335 | 1.7144 | -0.0233 |
| | | FTTransformer | 1.2922 | 0.9603 | -0.0024 | 0.7068 | 0.6176 | -0.005 | 4.851 | 1.7673 | -0.1635 |
| | | TabTransformer | 1.3326 | 0.9757 | 0.018 | 0.5687 | 0.6042 | 0.0144 | 4.4615 | 1.6531 | -0.0033 |
| | | DeepEnsemble | 0.7933 | 0.7212 | 0.456 | 0.5574 | 0.5875 | 0.1845 | 4.1893 | 1.4742 | 0.0191 |
| ViT-B | ✗ | BNN | 2.6959 | 1.3496 | -0.7066 | 235.1204 | 15.2681 | -399.8334 | 109.7285 | 10.2387 | -21.8722 |
| | | Attention | 0.3522 | 0.4541 | 0.7698 | 0.6386 | 0.6137 | -0.1672 | 5.4551 | 1.8235 | 0.0435 |
| | | Transformer | 0.7919 | 0.6919 | 0.4521 | 0.8142 | 0.6746 | -0.0291 | 4.5156 | 1.6711 | -0.0044 |
| | | iTransformer | 1.7446 | 1.0755 | 0.0012 | 0.7619 | 0.6755 | -0.0928 | 5.6567 | 1.8485 | -0.1427 |
| | | Bert | 1.3677 | 0.9535 | -0.0162 | 0.6175 | 0.5649 | -0.0177 | 3.9568 | 1.5058 | -0.0061 |
| | | FTTransformer | 1.5989 | 1.0241 | -0.0007 | 0.6935 | 0.6416 | 0 | 4.9066 | 1.8175 | -0.1035 |
| | | TabTransformer | 1.5013 | 0.9979 | -0.0087 | 0.5499 | 0.5782 | -0.0107 | 4.5243 | 1.6939 | -0.0004 |
| | | DeepEnsemble | 0.3404 | 0.3997 | 0.7865 | 0.5717 | 0.5784 | 0.0494 | 4.5123 | 1.6242 | 0.1628 |
| | ✓ | BNN | 2987.1626 | 48.6084 | -1996.2823 | 3488.9905 | 52.3622 | -6556.8347 | 2202.0063 | 40.3073 | -438.8903 |
| | | Attention | 0.8282 | 0.7311 | 0.4486 | 0.7802 | 0.6791 | -0.1223 | 5.1424 | 1.7877 | -0.0877 |
| | | Transformer | 0.6719 | 0.6726 | 0.4867 | 0.6051 | 0.6004 | -0.1034 | 5.5948 | 1.9038 | 0.0081 |
| | | iTransformer | 0.9682 | 0.7736 | 0.4246 | 0.6226 | 0.652 | -0.0676 | 4.2709 | 1.5372 | -0.0168 |
| | | Bert | 1.5141 | 1.0187 | -0.0194 | 0.6909 | 0.6452 | -0.0058 | 4.8081 | 1.6627 | -0.041 |
| | | FTTransformer | 1.5623 | 1.0412 | -0.0175 | 1.0005 | 0.7249 | -0.0218 | 4.2882 | 1.6657 | -0.1329 |
| | | TabTransformer | 1.3797 | 0.9389 | -0.0231 | 0.5677 | 0.5774 | -0.0278 | 4.4466 | 1.6685 | -0.0394 |
| | | DeepEnsemble | 0.9721 | 0.7553 | 0.3767 | 0.6961 | 0.6627 | -0.0353 | 4.7182 | 1.6365 | 0.0019 |
| | | BNN | 61.7642 | 7.4403 | -40.2051 | 69.1957 | 6.1824 | -107.3329 | 306.5371 | 17.3505 | -54.3146 |
| | | Attention | 1.6447 | 1.0194 | -0.0859 | 0.7738 | 0.708 | -0.1271 | 6.4196 | 1.9574 | -0.1871 |
| | | Transformer | 1.6154 | 1.026 | -0.0036 | 0.4828 | 0.5525 | -0.0215 | 5.7548 | 1.8716 | 0 |
| ViT-L | ✗ | iTransformer | 1.5482 | 1.035 | -0.0081 | 0.5705 | 0.5892 | -0.0617 | 4.517 | 1.5996 | -0.0285 |
| | | Bert | 1.4763 | 1.0169 | -0.0018 | 0.703 | 0.6155 | -0.1575 | 5.5751 | 1.7868 | -0.0479 |
| | | TabTransformer | 1.4291 | 0.9789 | -0.0083 | 0.6662 | 0.6251 | -0.0877 | 4.7353 | 1.7847 | -0.0759 |
| | | DeepEnsemble | 1.4613 | 0.9834 | -0.0015 | 0.5154 | 0.5475 | 0.2069 | 4.3184 | 1.5787 | 0.0011 |
| | ✓ | BNN | 1599.5596 | 36.275 | -1268.5395 | 722.1287 | 20.7319 | -1372.2653 | 3164.177 | 54.3953 | -736.5339 |
| | | Attention | 0.494 | 0.5728 | 0.6713 | 0.4823 | 0.5311 | 0.2585 | 3.8685 | 1.4727 | 0.1558 |
| | | Transformer | 0.7252 | 0.6709 | 0.5299 | 0.6652 | 0.6085 | -0.0026 | 4.5303 | 1.6469 | -0.0018 |
| | | iTransformer | 0.565 | 0.602 | 0.5652 | 0.8881 | 0.7227 | -0.1323 | 4.0502 | 1.6233 | -0.0326 |
| | | Bert | 1.4558 | 0.9899 | -0.0115 | 0.6754 | 0.6382 | -0.0262 | 4.7075 | 1.6946 | -0.0162 |
| | | FTTransformer | 1.5427 | 1.013 | -0.0375 | 0.5866 | 0.5586 | -0.0141 | 6.3804 | 2.0561 | -0.1931 |
| | | TabTransformer | 1.363 | 0.9538 | 0.001 | 0.7449 | 0.6417 | -0.0191 | 4.1693 | 1.6496 | 0.0132 |
| | | DeepEnsemble | 0.6889 | 0.6318 | 0.5491 | 0.4125 | 0.5127 | 0.3415 | 2.7836 | 1.305 | 0.3214 |
| DINOv2 ViT-S-Reg | ✗ | BNN | 22.6421 | 2.6388 | -15.5012 | 86.5596 | 9.1579 | -130.3714 | 223.8573 | 14.7969 | -52.0753 |
| | | Attention | 1.0694 | 0.7945 | 0.1487 | 0.4352 | 0.4927 | -0.0014 | 4.8158 | 1.7054 | -0.0018 |
| | | Transformer | 1.5397 | 0.9859 | -0.0002 | 0.7967 | 0.6928 | -0.0308 | 4.2255 | 1.585 | -0.0473 |
| | | iTransformer | 1.4849 | 0.9713 | -0.0005 | 0.8236 | 0.7016 | -0.166 | 4.3754 | 1.5781 | -0.0217 |
| | | Bert | 1.5523 | 1.0143 | -0.0017 | 0.8906 | 0.6829 | -0.1221 | 4.7521 | 1.6683 | -0.0164 |
| | | FTTransformer | 1.7299 | 1.1284 | -0.0059 | 0.5012 | 0.5564 | -0.0017 | 5.9356 | 1.9516 | -0.0659 |
| | | TabTransformer | 1.5322 | 1.0215 | -0.001 | 0.6485 | 0.6042 | -0.0115 | 5.2664 | 1.6837 | -0.0049 |
| | | DeepEnsemble | 1.2823 | 0.8719 | 0.1575 | 0.9132 | 0.7409 | -0.2041 | 5.723 | 1.9149 | -0.1614 |
| | ✓ | BNN | 320.0132 | 14.2083 | -249.206 | 223.7273 | 11.8503 | -373.0736 | 189.6176 | 11.227 | -30.7265 |
| | | Attention | 0.8129 | 0.6979 | 0.4854 | 0.5748 | 0.5816 | 0.1867 | 3.8498 | 1.4848 | 0.1042 |
| | | Transformer | 0.7351 | 0.6598 | 0.4646 | 0.5134 | 0.5538 | 0.2151 | 4.0783 | 1.5851 | 0.0084 |
| | | iTransformer | 0.8765 | 0.7015 | 0.3979 | 0.6157 | 0.5645 | 0.2138 | 4.3315 | 1.5713 | -0.0515 |
| | | Bert | 1.6669 | 1.069 | -0.0102 | 0.6118 | 0.6007 | -0.1107 | 4.7264 | 1.7177 | -0.0552 |
| | | FTTransformer | 1.4391 | 0.9956 | -0.0024 | 0.6098 | 0.5651 | -0.0249 | 6.0359 | 2.048 | -0.344 |
| | | TabTransformer | 1.5275 | 1.0281 | -0.0013 | 0.6597 | 0.6096 | 0.0039 | 3.6421 | 1.553 | -0.0204 |
| | | DeepEnsemble | 0.8935 | 0.7238 | 0.3797 | 0.5313 | 0.5475 | 0.2548 | 4.0215 | 1.5863 | -0.1102 |
| DINOv2 ViT-B-Reg | ✗ | BNN | 121.4875 | 10.5701 | -95.2073 | 30.9113 | 4.3105 | -48.3623 | 87.5109 | 9.0669 | -15.4107 |
| | | Attention | 0.9618 | 0.7795 | 0.3571 | 0.6463 | 0.6139 | -0.023 | 5.5201 | 1.7815 | -0.0752 |
| | | Transformer | 1.6544 | 1.058 | 0.0153 | 0.6403 | 0.5896 | -0.0495 | 4.8847 | 1.7828 | -0.0433 |
| | | iTransformer | 1.4031 | 0.9496 | -0.0186 | 0.904 | 0.7221 | -0.1267 | 6.2372 | 1.8708 | -0.0747 |
| | | Bert | 1.7473 | 1.1161 | -0.0093 | 0.8267 | 0.7103 | -0.4011 | 4.3127 | 1.6915 | -0.0001 |
| | | FTTransformer | 1.5192 | 0.9947 | -0.0002 | 0.612 | 0.5893 | -0.014 | 5.2733 | 1.8379 | -0.1447 |
| | | TabTransformer | 1.3473 | 0.9519 | -0.0081 | 0.6729 | 0.6435 | -0.0533 | 4.1648 | 1.5896 | -0.0003 |
| | | DeepEnsemble | 1.2766 | 0.8779 | 0.1481 | 0.6063 | 0.6209 | -0.0095 | 4.6372 | 1.6544 | -0.039 |
| | ✓ | BNN | 532.1342 | 20.5192 | -335.5921 | 345.3025 | 14.7426 | -654.807 | 252.9025 | 12.5931 | -48.2711 |
| | | Attention | 0.5821 | 0.5942 | 0.5741 | 0.496 | 0.561 | 0.2288 | 3.8 | 1.5215 | 0.2024 |
| | | Transformer | 0.6247 | 0.6514 | 0.5948 | 0.5683 | 0.5851 | 0.1903 | 5.0512 | 1.7242 | -0.0216 |
| | | iTransformer | 0.6074 | 0.603 | 0.5841 | 0.7568 | 0.6785 | -0.1097 | 4.9799 | 1.6511 | -0.0536 |
| | | Bert | 1.4708 | 1.0024 | 0 | 0.5757 | 0.5969 | -0.0115 | 4.6161 | 1.6495 | -0.035 |
| | | FTTransformer | 1.6628 | 1.0579 | -0.0159 | 0.4999 | 0.563 | -0.0066 | 6.2379 | 2.0213 | -0.1558 |
| | | TabTransformer | 1.4035 | 0.9672 | -0.0718 | 0.6917 | 0.6278 | 0.0062 | 5.2613 | 1.7857 | 0.0038 |
| | | DeepEnsemble | 0.4929 | 0.5469 | 0.5784 | 0.5249 | 0.5442 | 0.2792 | 3.574 | 1.4821 | 0.197 |
| DINOv2 ViT-L-Reg | ✓ | BNN | 499.9842 | 18.3083 | -320.3572 | 535.3407 | 20.306 | -908.5379 | 2805.533 | 50.8658 | -647.1655 |
| | | Attention | 0.5529 | 0.57 | 0.5786 | 0.5591 | 0.585 | 0.2352 | 5.2012 | 1.6832 | 0.0073 |
| | | Transformer | 0.6977 | 0.6425 | 0.5878 | 0.5901 | 0.5827 | -0.0131 | 4.0014 | 1.643 | -0.0158 |
| | | iTransformer | 0.6147 | 0.6 | 0.5983 | 0.5526 | 0.5928 | 0.1328 | 5.3984 | 1.7389 | -0.0452 |
| | | Bert | 1.4679 | 0.9858 | -0.0178 | 0.4567 | 0.5639 | 0.0023 | 4.5573 | 1.6165 | -0.0174 |
| | | FTTransformer | 1.474 | 1.0105 | -0.0125 | 0.7553 | 0.623 | -0.0129 | 5.6792 | 1.8972 | -0.1255 |
| | | TabTransformer | 1.7087 | 1.0703 | -0.0001 | 0.6213 | 0.6275 | -0.0632 | 4.6785 | 1.6617 | -0.0105 |
| | | DeepEnsemble | 0.5229 | 0.5572 | 0.6297 | 0.4283 | 0.4802 | 0.2887 | 3.5649 | 1.4699 | 0.0869 |

Table 5-3. Assessment of Regressor Heads for MRD1, MRD2, and LF tasks.

| Model | #Param | Task | Loss | OR | MRD1 | | | MRD2 | | | LF | | |
|---|---|---|---|---|---|---|---|---|---|---|---|---|---|
| | | | | | MSE | MAE | $R^2$ | MSE | MAE | $R^2$ | MSE | MAE | $R^2$ |
| ResNet18 | | Regression | MSE | ✗ | 0.2085 | 0.3465 | 0.8571 | 0.3715 | 0.4275 | 0.4933 | 2.3821 | 1.1888 | 0.4601 |
| | | | | ✓ | 0.2052 | 0.3332 | 0.8527 | 0.2685 | 0.3978 | 0.5163 | 2.721 | 1.2639 | 0.3592 |
| | | | Focal 0 | ✗ | 0.2033 | 0.3175 | 0.837 | 0.2872 | 0.434 | 0.4793 | **2.0514** | 1.1419 | 0.5842 |
| | | | | ✓ | 0.2279 | 0.3721 | 0.8524 | **0.2439** | 0.3805 | 0.6283 | 2.8805 | 1.3251 | 0.4846 |
| | | | Focal 2 | ✗ | 0.239 | 0.3693 | 0.8376 | 0.3082 | 0.435 | 0.4665 | 2.6267 | 1.2179 | 0.3829 |
| | | | | ✓ | **0.1785** | 0.3207 | 0.8883 | 0.254 | 0.4052 | 0.5058 | 2.1885 | 1.1462 | 0.2937 |
| | | | Focal 4 | ✗ | 0.203 | 0.3319 | 0.8467 | 0.3727 | 0.4388 | 0.4089 | 3.2122 | 1.3925 | 0.4231 |
| | | | | ✓ | 0.2441 | 0.378 | 0.8268 | 0.2679 | 0.3856 | 0.4638 | 2.722 | 1.2952 | 0.3576 |
| | | | Focal 6 | ✗ | 0.2336 | 0.3711 | 0.846 | 0.2672 | 0.4058 | 0.5214 | 2.2041 | 1.1734 | 0.5715 |
| | | | | ✓ | 0.304 | 0.3929 | 0.8029 | 0.3393 | 0.4158 | 0.518 | 3.0805 | 1.3699 | 0.4328 |
| | | Classification | BCE | ✗ | 1.1056 | 0.7846 | 0.3547 | 0.5826 | 0.5762 | 0.2813 | 5.4523 | 1.8195 | -0.6394 |
| | | | | ✓ | 0.7539 | 0.6823 | 0.4229 | 0.5847 | 0.5501 | 0.1404 | 6.7347 | 1.8996 | -0.3589 |
| | | | Focal 0 | ✗ | 0.8539 | 0.7246 | 0.4017 | 1.2439 | 0.8799 | -0.9529 | 7.0165 | 2.0195 | -1.0599 |
| | | | | ✓ | 1.1906 | 0.8045 | 0.2447 | 0.885 | 0.6758 | -0.0317 | 4.9605 | 1.7286 | 0.1433 |
| | | | Focal 2 | ✗ | 0.925 | 0.7263 | 0.3495 | 0.639 | 0.5841 | 0.2157 | 6.6139 | 2.0231 | -0.4448 |
| | | | | ✓ | 0.6903 | 0.6477 | 0.475 | 0.5492 | 0.5516 | 0.0984 | 3.9198 | 1.5203 | 0.0321 |
| | | | Focal 4 | ✗ | 0.7637 | 0.6924 | 0.462 | 0.5958 | 0.5371 | 0.2393 | 5.1167 | 1.7599 | 0.0271 |
| | | | | ✓ | 0.4079 | 0.5188 | 0.7186 | 0.4389 | 0.5107 | 0.261 | 3.3293 | 1.4144 | 0.3424 |
| | | | Focal 6 | ✗ | 0.8145 | 0.6913 | 0.4949 | 0.5196 | 0.5199 | 0.133 | 4.4804 | 1.6481 | -0.077 |
| | | | | ✓ | 0.4686 | 0.531 | 0.6641 | 0.8194 | 0.7023 | -0.0053 | 4.0475 | 1.6171 | 0.0768 |
| | | Regression | MSE | ✗ | **0.1711** | 0.3024 | 0.8773 | 0.2727 | 0.3664 | 0.6376 | 2.0693 | 1.1363 | 0.5749 |
| | | | | ✓ | 0.2157 | 0.3419 | 0.8613 | 0.3182 | 0.3818 | 0.5524 | 1.9198 | 1.017 | 0.5378 |
| | | | Focal 0 | ✗ | 0.226 | 0.3251 | 0.8389 | **0.2151** | 0.3563 | 0.5854 | 2.1648 | 1.1034 | 0.5411 |
| | | | | ✓ | 0.2544 | 0.3914 | 0.8253 | 0.3988 | 0.4178 | 0.4608 | 2.0052 | 1.0798 | 0.5763 |
| | | | Focal 2 | ✗ | 0.1715 | 0.3008 | 0.8741 | 0.2735 | 0.3895 | 0.561 | 2.1076 | 1.072 | 0.5331 |
| | | | | ✓ | 0.2665 | 0.3559 | 0.8029 | 0.331 | 0.3884 | 0.6006 | **1.7215** | 1.0274 | 0.5856 |

| | | | | | | | | | | | | | | |
|---|---|---|---|---|---|---|---|---|---|---|---|---|---|---|
| ResNet50 | | | Focal 4 | ✗ | 0.2249 | 0.3418 | 0.8682 | 0.3074 | 0.3893 | 0.5712 | 2.292 | 1.1602 | 0.53 |
| | | | | ✓ | 0.1741 | 0.3119 | 0.8805 | 0.3136 | 0.4025 | 0.5969 | 2.2415 | 1.1608 | 0.5786 |
| | | | Focal 6 | ✗ | 0.2284 | 0.3647 | 0.8559 | 0.2549 | 0.3808 | 0.535 | 1.7515 | 0.9568 | 0.6443 |
| | | | | ✓ | 0.2415 | 0.3613 | 0.827 | 0.32 | 0.4322 | 0.5034 | 2.0687 | 1.1105 | 0.5234 |
| | | Classification | BCE | ✗ | 0.5099 | 0.5484 | 0.6467 | 0.5395 | 0.5247 | 0.2206 | 4.0265 | 1.53 | -0.0848 |
| | | | | ✓ | 0.8519 | 0.6624 | 0.4416 | 0.6243 | 0.5384 | 0.0932 | 3.7606 | 1.438 | 0.1582 |
| | | | Focal 0 | ✗ | 0.8098 | 0.6746 | 0.3852 | 0.4209 | 0.4869 | 0.3057 | 4.9463 | 1.7994 | -0.2376 |
| | | | | ✓ | 1.2191 | 0.8102 | 0.3164 | 0.6737 | 0.5782 | 0.1661 | 5.9529 | 1.8786 | -0.4605 |
| | | | Focal 2 | ✗ | 0.6683 | 0.6516 | 0.5354 | 0.355 | 0.4632 | 0.3823 | 3.7326 | 1.3795 | 0.2445 |
| | | | | ✓ | 0.3941 | 0.4869 | 0.7623 | 0.4095 | 0.5022 | 0.2833 | 3.8707 | 1.5063 | 0.0722 |
| | | | Focal 4 | ✗ | 0.49 | 0.5521 | 0.6354 | 0.3729 | 0.4812 | 0.4231 | 6.0249 | 1.8578 | -0.4218 |
| | | | | ✓ | 0.7694 | 0.7115 | 0.5206 | 0.6212 | 0.6018 | 0.1485 | 3.0027 | 1.2876 | 0.3796 |
| | | | Focal 6 | ✗ | 0.6168 | 0.6344 | 0.5559 | 0.5983 | 0.6177 | -0.068 | 3.9125 | 1.5156 | 0.2068 |
| | | | | ✓ | 0.6875 | 0.6552 | 0.5433 | 0.7745 | 0.6843 | -0.3037 | 4.9095 | 1.7051 | 0.0008 |
| ResNet101 | | Regression | MSE | ✗ | 0.3654 | 0.4415 | 0.7706 | 0.3191 | 0.3836 | 0.5138 | 2.2094 | 1.1647 | 0.4631 |
| | | | | ✓ | 0.2216 | 0.3619 | 0.8657 | 0.3207 | 0.3959 | 0.5784 | **2.02** | 1.1388 | 0.603 |
| | | | Focal 0 | ✗ | 0.2567 | 0.3545 | 0.8114 | 0.2085 | 0.3629 | 0.6478 | 2.2284 | 1.1109 | 0.527 |
| | | | | ✓ | 0.2144 | 0.3594 | 0.8453 | 0.2565 | 0.3754 | 0.5693 | 2.2605 | 1.2196 | 0.4987 |
| | | | Focal 2 | ✗ | **0.1439** | 0.296 | 0.8807 | 0.2451 | 0.3933 | 0.6297 | 2.0991 | 1.0805 | 0.5466 |
| | | | | ✓ | 0.1682 | 0.3224 | 0.872 | 0.228 | 0.3728 | 0.6432 | 2.0603 | 1.1361 | 0.5953 |
| | | | Focal 4 | ✗ | 0.1675 | 0.3081 | 0.8808 | 0.6261 | 0.6531 | -0.1891 | 2.228 | 1.1562 | 0.5801 |
| | | | | ✓ | 0.158 | 0.3089 | 0.8903 | **0.1959** | 0.3322 | 0.6084 | 2.4979 | 1.2573 | 0.5813 |
| | | | Focal 6 | ✗ | 0.2845 | 0.3951 | 0.8065 | 0.416 | 0.4341 | 0.4932 | 2.0507 | 1.0807 | 0.5468 |
| | | | | ✓ | 0.2472 | 0.3618 | 0.8461 | 0.3497 | 0.4032 | 0.524 | 2.3083 | 1.1867 | 0.5201 |
| | | | BCE | ✗ | 0.6324 | 0.5979 | 0.5722 | 0.3512 | 0.4405 | 0.3476 | 5.9781 | 1.778 | -0.5147 |
| | | | | ✓ | 0.5697 | 0.5609 | 0.4973 | 0.445 | 0.5262 | 0.0997 | 3.841 | 1.482 | 0.0491 |
| | | | Focal 0 | ✗ | 0.5775 | 0.6059 | 0.5891 | 0.5684 | 0.5644 | 0.1092 | 3.9252 | 1.6101 | -0.0848 |
| | | | | ✓ | 0.895 | 0.6937 | 0.5104 | 0.6566 | 0.6395 | -0.1327 | 3.8074 | 1.5381 | 0.0548 |
| | | Classification | Focal 2 | ✗ | 0.5069 | 0.5526 | 0.5952 | 0.477 | 0.516 | 0.211 | 4.7014 | 1.6796 | -0.2866 |
| | | | | ✓ | 0.7505 | 0.6931 | 0.4769 | 0.4996 | 0.5188 | 0.1059 | 3.7262 | 1.4722 | 0.2287 |
| | | | Focal 4 | ✗ | 0.5048 | 0.5234 | 0.6151 | 0.4954 | 0.5546 | 0.0514 | 5.2471 | 1.8291 | -0.1578 |
| | | | | ✓ | 0.3682 | 0.4589 | 0.7625 | 0.5214 | 0.5816 | 0.1674 | 5.0176 | 1.7981 | 0.0218 |
| | | | Focal 6 | ✗ | 0.5394 | 0.5758 | 0.6459 | 0.3936 | 0.47 | 0.1576 | 5.1489 | 1.8285 | -0.0214 |
| | | | | ✓ | 0.7885 | 0.698 | 0.4795 | 0.5208 | 0.5604 | 0.1977 | 5.4328 | 1.9331 | 0.071 |
| EfficientNet B0 | 5M | Regression | MSE | ✗ | 0.1897 | 0.3342 | 0.8742 | 0.2379 | 0.3687 | 0.5412 | 2.4357 | 1.2333 | 0.4341 |
| | | | | ✓ | 0.1639 | 0.3057 | 0.8838 | 0.2715 | 0.3995 | 0.6247 | 3.0095 | 1.368 | 0.2349 |
| | | | Focal 0 | ✗ | 0.208 | 0.3414 | 0.8714 | 0.3732 | 0.4412 | 0.4681 | 3.1966 | 1.3729 | 0.2779 |
| | | | | ✓ | **0.1389** | **0.2943** | 0.9151 | 0.3528 | 0.4537 | 0.4234 | 2.2801 | 1.1958 | 0.5002 |
| | | | Focal 2 | ✗ | 0.1892 | 0.3385 | 0.8811 | **0.2032** | **0.3416** | 0.5804 | 2.5164 | 1.2423 | 0.4457 |
| | | | | ✓ | 0.304 | 0.4369 | 0.745 | 0.3239 | 0.3977 | 0.571 | 2.4259 | 1.2237 | 0.4961 |
| | | | Focal 4 | ✗ | 0.2252 | 0.3712 | 0.8665 | 0.3008 | 0.4115 | 0.6098 | 1.9471 | 1.0928 | 0.4739 |
| | | | | ✓ | 0.356 | 0.4649 | 0.7607 | 0.3435 | 0.4605 | 0.4624 | 2.3912 | 1.1953 | 0.5116 |
| | | | Focal 6 | ✗ | 0.197 | 0.3449 | 0.87 | 0.2634 | 0.395 | 0.5754 | **1.8112** | **1.0657** | 0.5842 |
| | | | | ✓ | 0.3094 | 0.4313 | 0.7697 | 1.0536 | 0.8787 | -0.653 | 2.8954 | 1.3041 | 0.4384 |
| | | Classification | BCE | ✗ | 0.3233 | 0.4382 | 0.7537 | 0.3857 | 0.4688 | 0.1828 | 6.5635 | 1.9495 | -0.6147 |
| | | | | ✓ | 0.5981 | 0.5694 | 0.5805 | 0.5379 | 0.5343 | 0.087 | 6.3164 | 1.9114 | -0.144 |
| | | | Focal 0 | ✗ | 0.7314 | 0.6441 | 0.5013 | 0.5336 | 0.567 | 0.2545 | 6.5553 | 1.9428 | -0.3516 |
| | | | | ✓ | 1.3731 | 0.976 | 0.0329 | 0.5223 | 0.5629 | 0.2333 | 5.0322 | 1.796 | 0.0511 |
| | | | Focal 2 | ✗ | 0.5199 | 0.5354 | 0.5784 | 0.657 | 0.614 | -0.1873 | 4.3896 | 1.7061 | -0.0473 |
| | | | | ✓ | 0.5037 | 0.5556 | 0.6224 | 0.5612 | 0.5666 | 0.2624 | 6.3694 | 2.015 | -0.272 |
| | | | Focal 4 | ✗ | 0.667 | 0.6388 | 0.6301 | 0.4179 | 0.4925 | 0.2027 | 5.408 | 1.8577 | -0.0164 |
| | | | | ✓ | 0.7273 | 0.6801 | 0.4523 | 0.6423 | 0.6385 | -0.0067 | 5.2251 | 1.7281 | -0.1358 |
| | | | Focal 6 | ✗ | 0.6989 | 0.6133 | 0.3881 | 0.94 | 0.7247 | -0.2699 | 5.5268 | 1.8263 | -0.2096 |
| | | | | ✓ | 1.5501 | 1.0159 | -0.1172 | 0.7683 | 0.6684 | -0.1327 | 4.8471 | 1.6954 | -0.1695 |
| | | | MSE | ✗ | 0.158 | **0.291** | 0.9023 | **0.2467** | **0.3667** | 0.5694 | 6.9066 | 1.9821 | -0.6483 |
| | | | | ✓ | 0.3687 | 0.4762 | 0.7347 | 0.2645 | 0.3989 | 0.5746 | 2.6081 | 1.2713 | 0.338 |
| EfficientNet B4 | 19M | Regression | Focal 0 | ✗ | 0.1766 | 0.3114 | 0.855 | 0.3424 | 0.4215 | 0.4913 | 3.2187 | 1.3753 | 0.3151 |
| | | | | ✓ | 0.1935 | 0.3439 | 0.8549 | 0.3195 | 0.4126 | 0.5474 | 2.3156 | 1.2256 | 0.4552 |
| | | | Focal 2 | ✗ | 0.3659 | 0.4634 | 0.7558 | 0.3788 | 0.436 | 0.379 | 3.1056 | 1.3493 | 0.511 |
| | | | | ✓ | **0.1438** | 0.2978 | 0.9024 | 0.2951 | 0.3862 | 0.5106 | 2.7031 | 1.3046 | 0.3441 |
| | | | Focal 4 | ✗ | 9.0546 | 2.6465 | -5.169 | 0.2774 | 0.3822 | 0.5452 | **2.0575** | **1.0941** | 0.5151 |
| | | | | ✓ | 0.1799 | 0.3223 | 0.8599 | 0.7311 | 0.6861 | -0.0204 | 2.2271 | 1.1963 | 0.4884 |
| | | | Focal 6 | ✗ | 0.2994 | 0.418 | 0.8124 | 0.7471 | 0.5867 | -0.0009 | 2.6865 | 1.2685 | 0.4581 |
| | | | | ✓ | 0.2294 | 0.3595 | 0.855 | 0.4 | 0.4551 | 0.5227 | 3.2216 | 1.3935 | 0.2955 |
| | | Classification | BCE | ✗ | 0.6955 | 0.6423 | 0.5297 | 0.5396 | 0.5681 | 0.1775 | 8.3628 | 2.3287 | -0.535 |
| | | | | ✓ | 0.4058 | 0.4823 | 0.7048 | 0.4391 | 0.4707 | 0.257 | 7.4886 | 2.0639 | -1.055 |
| | | | Focal 0 | ✗ | 0.6005 | 0.5753 | 0.5779 | 0.6826 | 0.6212 | -0.2869 | 4.0057 | 1.5942 | -0.2667 |
| | | | | ✓ | 0.6466 | 0.6218 | 0.543 | 1.0108 | 0.8027 | -0.4734 | 6.3324 | 2.013 | -0.4419 |
| | | | Focal 2 | ✗ | 0.6354 | 0.6316 | 0.6126 | 0.3181 | 0.4529 | 0.3385 | 5.6923 | 1.8935 | -0.4046 |
| | | | | ✓ | 1.0268 | 0.8208 | 0.3238 | 0.836 | 0.7503 | -0.4861 | 8.4983 | 2.468 | -0.9251 |
| | | | Focal 4 | ✗ | 0.5739 | 0.579 | 0.6128 | 0.583 | 0.5969 | 0.0818 | 4.1301 | 1.5632 | -0.0203 |
| | | | | ✓ | 0.7526 | 0.6796 | 0.4968 | 0.4441 | 0.5309 | -0.0353 | 5.1468 | 1.7365 | -0.0377 |
| | | | Focal 6 | ✗ | 1.1399 | 0.8248 | 0.2949 | 0.9266 | 0.7221 | -0.1842 | 5.3842 | 1.7344 | -0.1207 |
| | | | | ✓ | 1.8145 | 1.0781 | -0.0833 | 0.9962 | 0.7263 | -0.2278 | 4.8209 | 1.735 | -0.4903 |
| EfficientNet | 66M | Regression | MSE | ✗ | 0.2537 | 0.3562 | 0.852 | 0.3715 | 0.4486 | 0.2819 | | 2.6326 | -6.9898 |
| | | | | ✓ | 0.2588 | 0.3502 | 0.8371 | 153.4357 | 6.3873 | -248.5624 | | 32.4892 | -2597.796 |
| | | | Focal 0 | ✗ | **0.1366** | **0.2638** | 0.8884 | 0.7491 | 0.4755 | -0.1112 | | 1.6117 | -0.0947 |
| | | | | ✓ | 0.1479 | 0.2925 | 0.889 | 94461.62 | 379.6557 | 42134.70 | | 25.0351 | 3357.225 |
| | | | Focal 2 | ✗ | 0.4465 | 0.4897 | 0.6859 | 3.2087 | 0.853 | -4.4967 | | 1.3158 | 0.4967 |
| | | | | ✓ | 0.2814 | 0.3884 | 0.8057 | 0.3832 | 0.5189 | 0.3183 | | **1.3074** | 0.3897 |
| | | | Focal 4 | ✗ | 0.2785 | 0.421 | 0.8101 | 0.436 | 0.5385 | 0.2742 | | 1.3504 | 0.3318 |
| | | | | ✓ | 0.1985 | 0.3321 | 0.873 | **0.2578** | **0.3869** | 0.4554 | | 1.3522 | 0.2334 |
| | | | Focal 6 | ✗ | 0.2292 | 0.3567 | 0.8257 | 0.6985 | 0.6573 | -0.1202 | | 1.4528 | 0.2422 |
| | | | | ✓ | 0.2006 | 0.3302 | 0.8456 | 0.4431 | 0.5378 | 0.1671 | | 1.5295 | 0.2412 |
| et B7 | 66M | Classification | BCE | ✗ | 1.7419 | 1.0921 | -0.0037 | 0.722 | 0.6124 | -0.0044 | 3.6975 | 1.4788 | 0.0027 |
| | | | | ✓ | 1.5953 | 1.0255 | -0.0023 | 0.5722 | 0.5866 | -0.033 | 14.6046 | 2.7275 | -2.4367 |
| | | | Focal 0 | ✗ | 0.7548 | 0.6322 | 0.3916 | 2.1548 | 1.3072 | -2.6876 | 15.7706 | 3.5908 | -3.0404 |
| | | | | ✓ | 0.7047 | 0.613 | 0.5696 | 3.9643 | 1.5023 | -4.983 | 16.1627 | 3.4793 | -2.2642 |
| | | | Focal 2 | ✗ | 1.2881 | 0.9439 | 0.2096 | 1.0977 | 0.8986 | -1.0834 | 7.3668 | 2.1774 | -0.3418 |
| | | | | ✓ | 0.5325 | 0.5516 | 0.6175 | 1.1899 | 0.9212 | -1.2469 | 6.1983 | 2.0375 | -0.2482 |
| | | | Focal 4 | ✗ | 1.603 | 1.0234 | -0.0848 | 0.7887 | 0.6817 | -0.2834 | 4.8191 | 1.615 | -0.0418 |
| | | | | ✓ | 1.5015 | 0.9875 | -0.0077 | 0.6186 | 0.6215 | -0.0001 | 3.8526 | 1.5818 | 0.0131 |
| | | | Focal 6 | ✗ | 0.7969 | 0.7343 | 0.4237 | 0.8674 | 0.7329 | -0.4242 | 6.6705 | 2.0484 | -0.1694 |
| | | | | ✓ | 1.8005 | 1.0769 | -0.1894 | 0.8738 | 0.7204 | -0.4054 | 4.7519 | 1.7451 | -0.319 |
| ViT-T | | Regression | MSE | ✗ | 0.3374 | 0.421 | 0.7708 | 0.3899 | 0.456 | 0.4545 | 4.9945 | 1.8719 | -0.1593 |
| | | | | ✓ | 0.2328 | 0.3981 | 0.7804 | 0.33 | 0.4074 | 0.4251 | 4.0816 | 1.5446 | 0.0905 |
| | | | Focal 0 | ✗ | 0.3776 | 0.4702 | 0.7607 | 0.8064 | 0.6895 | -0.0614 | 4.2446 | 1.6944 | 0.1585 |
| | | | | ✓ | 0.3256 | 0.428 | 0.726 | 0.7438 | 0.6968 | -0.2257 | **3.2322** | 1.4573 | 0.2131 |
| | | | Focal 2 | ✗ | **0.235** | 0.3697 | 0.8374 | 0.3367 | 0.4575 | 0.4619 | 4.1101 | 1.5777 | -0.0865 |
| | | | | ✓ | 0.3319 | 0.4318 | 0.778 | 0.357 | 0.4548 | 0.3595 | 5.5936 | 1.8828 | -0.5937 |
| | | | Focal 4 | ✗ | 0.3843 | 0.4913 | 0.7345 | 0.7363 | 0.6456 | 0.0697 | 4.2777 | 1.6484 | -0.0176 |
| | | | | ✓ | 0.3672 | 0.4577 | 0.7734 | 0.5649 | 0.5633 | 0.211 | 5.3006 | 1.7921 | -0.2892 |
| | | | Focal 6 | ✗ | 0.2867 | 0.4191 | 0.8127 | 0.3465 | 0.4377 | 0.3957 | 4.998 | 1.7998 | -0.4005 |
| | | | | ✓ | 0.5195 | 0.5459 | 0.6547 | **0.3052** | 0.4231 | 0.4378 | 4.4739 | 1.6675 | -0.0921 |
| | | Classification | BCE | ✗ | 0.6646 | 0.6446 | 0.5912 | 0.8647 | 0.7443 | -0.2543 | 5.4387 | 1.9077 | -0.5211 |
| | | | | ✓ | 0.7894 | 0.6845 | 0.4747 | 0.6456 | 0.6201 | -0.0364 | 6.8327 | 2.0461 | -0.4762 |
| | | | Focal 0 | ✗ | 1.0821 | 0.715 | 0.2557 | 0.8018 | 0.7172 | -0.0425 | 5.9909 | 1.872 | -0.4106 |
| | | | | ✓ | 0.7309 | 0.6285 | 0.5746 | 0.5786 | 0.5888 | 0.0362 | 8.6678 | 2.3177 | -1.1405 |
| | | | Focal 2 | ✗ | 1.1611 | 0.867 | 0.2166 | 0.4911 | 0.5347 | 0.326 | 8.6018 | 2.2858 | -0.4612 |
| | | | | ✓ | 0.8948 | 0.7292 | 0.4108 | 0.4892 | 0.526 | 0.2962 | 6.5529 | 2.0195 | -0.4307 |
| | | | Focal 4 | ✗ | 0.5261 | 0.5524 | 0.6715 | 0.705 | 0.6592 | -0.3768 | 5.6211 | 1.8656 | -0.3305 |
| | | | | ✓ | 0.4775 | 0.5306 | 0.615 | 0.7729 | 0.6799 | -0.0968 | 7.6943 | 2.0968 | -0.4079 |
| | | | Focal 6 | ✗ | 0.6023 | 0.6113 | 0.6606 | 1.2168 | 0.8356 | -0.7386 | 4.5117 | 1.6903 | 0.0999 |
| | | | | ✓ | 1.1513 | 0.8259 | 0.2463 | 0.4061 | 0.5179 | 0.2599 | 5.313 | 1.8252 | -0.1504 |
| ViT-S | | Regression | MSE | ✗ | 0.3449 | 0.4238 | 0.8037 | 0.3665 | 0.4605 | 0.5476 | **2.9381** | 1.3451 | 0.3889 |
| | | | | ✓ | 0.2293 | 0.3688 | 0.8532 | 0.3795 | 0.4848 | 0.5365 | 3.1046 | 1.448 | 0.3233 |
| | | | Focal 0 | ✗ | 0.1989 | 0.3223 | 0.8517 | 0.3835 | 0.4659 | 0.4079 | 3.5816 | 1.5202 | 0.0759 |
| | | | | ✓ | 0.2605 | 0.3949 | 0.7979 | 0.2935 | 0.4329 | 0.5679 | 4.1337 | 1.5983 | -0.0285 |
| | | | Focal 2 | ✗ | **0.1987** | 0.3626 | 0.8828 | 0.3399 | 0.4282 | 0.5049 | 4.5228 | 1.6982 | -0.1991 |
| | | | | ✓ | 0.24 | 0.3731 | 0.8318 | **0.2542** | 0.3931 | 0.5975 | 5.2473 | 1.8487 | -0.1063 |
| | | | Focal 4 | ✗ | 0.3972 | 0.4695 | 0.7414 | 0.3602 | 0.4715 | 0.3768 | 5.3618 | 1.8393 | -0.095 |
| | | | | ✓ | 0.3379 | 0.4148 | 0.7742 | 0.2598 | 0.4003 | 0.5266 | 5.1837 | 1.776 | -0.1354 |
| | | | Focal 6 | ✗ | 0.3068 | 0.4121 | 0.8122 | 0.3999 | 0.4968 | 0.4148 | 3.8469 | 1.5206 | 0.2505 |
| | | | | ✓ | 0.298 | 0.3601 | 0.7918 | 0.2925 | 0.4299 | 0.3021 | 3.7845 | 1.5579 | 0.3273 |
| | | Classification | BCE | ✗ | 0.4489 | 0.5101 | 0.677 | 0.5452 | 0.5754 | -0.1275 | 3.9868 | 1.5609 | 0.013 |
| | | | | ✓ | 0.6799 | 0.6371 | 0.5296 | 0.6076 | 0.6033 | -0.2485 | 4.2436 | 1.5799 | -0.1733 |
| | | | Focal 0 | ✗ | 0.898 | 0.7364 | 0.2801 | 0.8699 | 0.7126 | -0.0219 | 7.5976 | 2.1493 | -0.6867 |
| | | | | ✓ | 1.0098 | 0.7475 | 0.1616 | 0.9418 | 0.7009 | -0.0527 | 7.4103 | 2.1491 | -0.4224 |
| | | | Focal 2 | ✗ | 0.7979 | 0.6725 | 0.5006 | 0.4747 | 0.5377 | 0.1636 | 5.356 | 1.7628 | -0.2837 |
| | | | | ✓ | 0.7155 | 0.645 | 0.5362 | 0.5333 | 0.5541 | 0.1893 | 8.3664 | 2.473 | -0.8323 |
| | | | Focal 4 | ✗ | 0.5731 | 0.6035 | 0.5699 | 0.5654 | 0.5425 | 0.1193 | 4.7935 | 1.7099 | -0.1813 |
| | | | | ✓ | 0.5767 | 0.6105 | 0.5763 | 0.546 | 0.5702 | -0.0065 | 5.1826 | 1.8245 | -0.1594 |
| | | | Focal 6 | ✗ | 0.5063 | 0.5508 | 0.6796 | 1.073 | 0.8187 | -0.4915 | 6.461 | 2.1117 | -0.4709 |
| | | | | ✓ | 0.7086 | 0.632 | 0.4785 | 0.4835 | 0.5204 | 0.1306 | 4.2248 | 1.687 | -0.0568 |
| | | Regression | MSE | ✗ | 0.2288 | 0.353 | 0.8127 | 0.2531 | 0.3937 | 0.5225 | 3.9332 | 1.5739 | 0.1192 |
| | | | | ✓ | 0.3073 | 0.4271 | 0.8215 | 0.3269 | 0.4506 | 0.5293 | 4.8499 | 1.7638 | 0.06 |
| | | | Focal 0 | ✗ | 0.2412 | 0.3616 | 0.8279 | 0.2359 | 0.37 | 0.6597 | 4.2851 | 1.6769 | 0.0367 |
| | | | | ✓ | 0.1897 | 0.3212 | 0.8663 | 0.2613 | 0.3745 | 0.5792 | 4.2023 | 1.6459 | 0.135 |
| | | | Focal 2 | ✗ | 0.2479 | 0.363 | 0.8315 | **0.1988** | 0.3383 | 0.646 | 3.996 | 1.5544 | 0.1376 |
| | | | | ✓ | 0.2179 | 0.3586 | 0.8684 | 0.4172 | 0.4988 | 0.4112 | 3.9151 | 1.6474 | 0.2993 |
| ViT-B | | | Focal 4 | ✗ | 0.3492 | 0.4021 | 0.7944 | 0.2787 | 0.4159 | 0.4454 | 4.9059 | 1.7892 | 0.0536 |
| | | | | ✓ | 0.2879 | 0.3959 | 0.8345 | 0.2981 | 0.4306 | 0.5401 | 3.6189 | 1.4844 | 0.1638 |
| | | | Focal 6 | ✗ | 0.2411 | 0.3946 | 0.8361 | 0.5705 | 0.5993 | -0.0955 | 4.2892 | 1.6512 | 0.0558 |
| | | | | ✓ | **0.1854** | 0.3316 | 0.8659 | 0.8475 | 0.6992 | -0.0366 | **2.9168** | 1.3724 | 0.2973 |
| | | Classification | BCE | ✗ | 0.5436 | 0.5753 | 0.6148 | 0.8581 | 0.6765 | -0.1845 | 5.5775 | 1.8786 | -0.2246 |
| | | | | ✓ | 0.5324 | 0.5814 | 0.6249 | 0.5682 | 0.5657 | 0.105 | 6.7712 | 2.0333 | -0.5114 |
| | | | Focal 0 | ✗ | 1.1055 | 0.7597 | 0.1801 | 0.8374 | 0.6991 | -0.5091 | 8.6239 | 2.3245 | -0.5222 |
| | | | | ✓ | 0.8019 | 0.7025 | 0.3582 | 0.4993 | 0.5744 | -0.1046 | 6.6843 | 1.9127 | -0.2652 |
| | | | Focal 2 | ✗ | 1.1162 | 0.8281 | 0.2096 | 0.4821 | 0.5391 | 0.1954 | 3.2434 | 1.3442 | 0.0766 |
| | | | | ✓ | 1.1529 | 0.8367 | 0.1526 | 0.5445 | 0.5433 | 0.1321 | 6.0496 | 1.873 | -0.224 |
| | | | Focal 4 | ✗ | 0.4788 | 0.5311 | 0.7022 | 0.34 | 0.4659 | 0.4092 | 3.9506 | 1.5383 | 0.0586 |
| | | | | ✓ | 0.6333 | 0.6061 | 0.6132 | 0.4343 | 0.4784 | 0.2153 | 6.6835 | 2.0688 | -0.1966 |
| | | | Focal 6 | ✗ | 0.3878 | 0.4817 | 0.7278 | 0.4812 | 0.528 | 0.244 | 4.4238 | 1.6138 | -0.0699 |
| | | | | ✓ | 0.4149 | 0.4975 | 0.7159 | 0.9045 | 0.7231 | -0.551 | 5.5688 | 1.935 | -0.5606 |
| DINOv2 ViT-S-Reg | 21M | Regression | MSE | ✗ | 1.2559 | 0.91 | -0.0073 | 0.7369 | 0.635 | -0.2276 | 4.1763 | 1.5925 | -0.0763 |
| | | | | ✓ | 1.6235 | 1.022 | -0.0096 | 0.5994 | **0.5661** | -0.0557 | 4.3649 | 1.6053 | -0.0037 |
| | | | Focal 0 | ✗ | 1.5557 | 1.0136 | -0.0231 | **0.533** | 0.5772 | -0.0345 | 4.7589 | 1.7167 | -0.016 |
| | | | | ✓ | 1.4923 | 0.9712 | -0.0013 | 0.7635 | 0.6787 | -0.0236 | 7.1988 | 2.0416 | -0.4027 |
| | | | Focal 2 | ✗ | 1.3212 | 0.938 | -0.0002 | 0.9032 | 0.7167 | -0.1607 | 4.5448 | 1.6527 | -0.0005 |
| | | | | ✓ | 1.7691 | 1.0584 | -0.0437 | 0.7343 | 0.6222 | -0.1363 | 4.0948 | 1.5662 | **-0.0002** |
| | | | Focal 4 | ✗ | 1.36 | 0.9496 | -0.0649 | 0.7654 | 0.6671 | -0.0818 | 4.1721 | 1.618 | -0.0149 |
| | | | | ✓ | 1.7252 | 1.078 | -0.0005 | 0.6673 | 0.6336 | -0.0647 | 4.8148 | 1.6939 | -0.0874 |
| | | | Focal 6 | ✗ | 1.7618 | 1.0794 | -0.0998 | 0.6254 | 0.6219 | -0.0002 | 4.9297 | 1.657 | -0.0278 |
| | | | | ✓ | 1.3165 | 0.9416 | -0.0009 | 0.7215 | 0.6536 | -0.0265 | 5.1131 | 1.7467 | -0.0204 |
| | | | BCE | ✗ | 1.4207 | 0.9709 | -0.0077 | 0.7458 | 0.6368 | -0.0005 | **3.4955** | **1.523** | -0.02 |
| | | | | ✓ | 1.6481 | 1.0474 | -0.0283 | 0.563 | 0.5779 | -0.0112 | 4.5607 | 1.6252 | -0.0652 |
| | | | Focal 0 | ✗ | 3.986 | 1.7208 | -1.7839 | 2.1277 | 1.3195 | -3.2963 | 16.1937 | 3.5841 | -2.6918 |
| | | | | ✓ | 3.342 | 1.5489 | -1.3632 | 2.2094 | 1.2989 | -1.6113 | 15.1326 | 3.4275 | -2.0773 |
| | | Classification | Focal 2 | ✗ | 1.322 | 0.9431 | -0.0441 | 1.3642 | 1.003 | -1.4051 | 6.6014 | 2.1492 | -0.5301 |
| | | | | ✓ | 1.7116 | 1.0931 | -0.0181 | 1.081 | 0.8719 | -0.7814 | 6.4167 | 2.0184 | -0.2189 |
| | | | Focal 4 | ✗ | **1.205** | **0.8844** | -0.0119 | 0.7446 | 0.6659 | -0.0014 | 4.7425 | 1.7345 | -0.0717 |
| | | | | ✓ | 1.4388 | 0.9642 | -0.0036 | 0.7077 | 0.6531 | -0.0161 | 5.5953 | 1.7804 | -0.0137 |
| | | | Focal 6 | ✗ | 1.4688 | 0.9574 | -0.0539 | 1.0387 | 0.8198 | -0.6339 | 5.7483 | 1.8684 | -0.2987 |
| | | | | ✓ | 1.9564 | 1.1379 | -0.1361 | 0.8705 | 0.6817 | -0.1762 | 4.67 | 1.7104 | -0.3238 |
| DINOv2 ViT-B-Reg | 86M | Regression | MSE | ✗ | 1.5842 | 1.0266 | 0 | 0.597 | 0.6111 | -0.0047 | 4.9982 | 1.7741 | -0.0432 |
| | | | | ✓ | 1.5114 | 1.0045 | -0.0241 | 0.9157 | 0.7507 | -0.3495 | 7.174 | 2.2465 | -0.5088 |
| | | | Focal 0 | ✗ | 1.4296 | 0.9622 | -0.0166 | 0.6609 | 0.6489 | -0.2696 | 5.5358 | 1.9651 | -0.2155 |
| | | | | ✓ | **1.2217** | 0.9047 | -0.0464 | **0.511** | 0.5777 | -0.0927 | 5.221 | 1.7009 | -0.0424 |
| | | | Focal 2 | ✗ | 1.6305 | 1.0679 | -0.0983 | 0.7857 | 0.7049 | -0.181 | 5.3721 | 1.8707 | -0.0246 |
| | | | | ✓ | 1.757 | 1.0705 | -0.1053 | 0.6302 | 0.6253 | -0.0692 | 5.1011 | 1.7347 | **-0.0014** |
| | | | Focal 4 | ✗ | 1.2391 | **0.9029** | -0.0007 | 0.7505 | 0.6774 | -0.1377 | 4.8326 | 1.8223 | -0.1207 |
| | | | | ✓ | 1.3839 | 0.9822 | -0.0882 | 0.6174 | 0.5794 | -0.0009 | 6.8238 | 2.1744 | -0.7644 |
| | | | Focal 6 | ✗ | 1.4934 | 1.0059 | -0.0178 | 0.5733 | **0.5741** | -0.035 | 4.9468 | 1.7802 | -0.1141 |
| | | | | ✓ | 1.521 | 0.9946 | -0.0307 | 0.7711 | 0.6569 | -0.0484 | 4.6587 | 1.6694 | -0.034 |
| | | Classification | BCE | ✗ | 1.539 | 1.0136 | -0.0015 | 0.6071 | 0.6136 | -0.0302 | **4.4622** | **1.7002** | -0.0255 |
| | | | | ✓ | 1.3509 | 0.9382 | -0.0002 | 0.8219 | 0.6838 | -0.0089 | 5.1347 | 1.7011 | -0.0237 |
| | | | Focal 0 | ✗ | 3.0346 | 1.4581 | -1.1709 | 2.3229 | 1.3053 | -2.093 | 16.2079 | 3.5772 | -2.0845 |
| | | | | ✓ | 2.9582 | 1.4567 | -1.2632 | 2.1768 | 1.2637 | -1.8805 | 17.1349 | 3.7006 | -2.8283 |
| | | | Focal 2 | ✗ | 1.569 | 1.0315 | -0.0371 | 1.0595 | 0.8545 | -0.952 | 5.9121 | 1.9797 | -0.3343 |
| | | | | ✓ | 1.5303 | 1.0209 | -0.0138 | 1.1226 | 0.882 | -0.9649 | 6.0034 | 1.9802 | -0.2851 |
| | | | Focal 4 | ✗ | 1.4758 | 0.9886 | -0.0231 | 0.6919 | 0.6009 | -0.1154 | 5.6915 | 1.7314 | -0.0076 |
| | | | | ✓ | 1.4524 | 0.9554 | -0.0422 | 0.6965 | 0.6387 | -0.0002 | 5.1091 | 1.702 | -0.0148 |
| | | | Focal 6 | ✗ | 1.7399 | 1.054 | -0.1379 | 0.995 | 0.7824 | -0.5009 | 6.4358 | 1.9392 | -0.2758 |
| | | | | ✓ | 1.5064 | 0.941 | -0.1784 | 0.6855 | 0.649 | -0.3247 | 6.2572 | 1.9014 | -0.3032 |
| | | | MSE | ✗ | 1.5432 | 1.0245 | 0.0772 | 0.5811 | 0.6061 | -0.0177 | 5.0566 | 1.787 | -0.1105 |
| | | | | ✓ | 1.3976 | 0.9661 | -0.0334 | 0.5679 | 0.5941 | -0.0444 | 4.7434 | 1.802 | -0.0532 |
| DINOv2 ViT-L-Reg | 300M | Regression | Focal 0 | ✗ | 1.4277 | 0.9911 | -0.0173 | 0.6589 | 0.6177 | -0.0774 | 5.4063 | 1.8818 | -0.2787 |
| | | | | ✓ | 1.4353 | 0.9405 | -0.0717 | 0.5841 | 0.6067 | -0.0026 | 4.3892 | 1.6581 | -0.1301 |
| | | | Focal 2 | ✗ | 1.5498 | 1.0056 | -0.0092 | 0.8025 | 0.6433 | -0.0053 | 4.8259 | 1.804 | -0.0749 |
| | | | | ✓ | 1.5569 | 1.0165 | -0.0972 | 0.6704 | 0.6641 | -0.1285 | 5.1592 | 1.7309 | -0.1561 |
| | | | Focal 4 | ✗ | 1.599 | 1.0454 | -0.0019 | 0.7303 | 0.6879 | -0.1474 | 4.8389 | 1.672 | -0.0242 |
| | | | | ✓ | 1.458 | 1.0037 | -0.0011 | 0.6481 | 0.6388 | -0.0019 | 5.5072 | 1.7878 | -0.0974 |
| | | | Focal 6 | ✗ | **1.1784** | **0.8848** | -0.0125 | 1.1896 | 0.8624 | -0.6412 | 5.0868 | 1.8563 | -0.2732 |
| | | | | ✓ | 1.5022 | 1.009 | -0.0023 | 0.5662 | 0.5336 | -0.0213 | 5.1308 | 1.7359 | -0.0106 |
| | | Classification | BCE | ✗ | 1.3241 | 0.9497 | -0.0006 | **0.3549** | **0.4671** | -0.0109 | **3.4516** | **1.524** | **-0.0054** |
| | | | | ✓ | 1.3868 | 0.9444 | -0.0103 | 0.5651 | 0.5936 | 0 | 4.8656 | 1.6875 | -0.0225 |
| | | | Focal 0 | ✗ | 3.5593 | 1.548 | -1.0386 | 1.8244 | 1.1921 | -1.9566 | 16.6402 | 3.6734 | -2.7825 |
| | | | | ✓ | 2.9547 | 1.4123 | -1.2606 | 2.4309 | 1.3936 | -2.4944 | 16.3777 | 3.609 | -2.0806 |
| | | | Focal 2 | ✗ | 1.9209 | 1.1313 | -0.0082 | 1.2332 | 0.9123 | -1.3529 | 6.7384 | 2.1105 | -0.5539 |
| | | | | ✓ | 1.2672 | 0.9374 | 0.111 | 1.1912 | 0.9085 | -1.0935 | 6.0772 | 1.9194 | -0.1134 |
| | | | Focal 4 | ✗ | 1.6546 | 1.0299 | -0.0179 | 0.5519 | 0.5588 | -0.0015 | 5.5488 | 1.797 | -0.0843 |
| | | | | ✓ | 1.4711 | 0.9713 | -0.0269 | 0.7005 | 0.6439 | -0.0075 | 3.9275 | 1.5404 | -0.0121 |
| | | | Focal 6 | ✗ | 1.6209 | 1.0201 | -0.093 | 0.7832 | 0.7116 | -0.3592 | 7.5904 | 2.0769 | -0.3214 |
| | | | | ✓ | 1.5206 | 0.9717 | -0.1644 | 0.7963 | 0.6664 | -0.1944 | 6.5569 | 1.962 | -0.2844 |

Table 5-4. Assessment of Binary Encoding, Focal Loss, and Orthogonal Regularization for MRD1, MRD2, and LF tasks. Classification means 16-bit binary encoding applied. Otherwise, Regression. The number following the “Focal” means the gamma value.

| Model | FPN | SSP | MRD1 | | | MRD2 | | | LF | | |
|---|---|---|---|---|---|---|---|---|---|---|---|
| | | | MSE | MAE | $R^2$ | MSE | MAE | $R^2$ | MSE | MAE | $R^2$ |
| ResNet18 | ✗ | ✗ | 0.2656 | 0.3944 | 0.8129 | 0.3304 | 0.436 | 0.5111 | 2.6213 | 1.2564 | 0.4471 |
| | | ✓ | 0.3252 | 0.4217 | 0.7652 | 0.2968 | 0.4094 | 0.5281 | 2.3417 | 1.2112 | 0.3088 |
| | ✓ | ✗ | 0.7238 | 0.6873 | 0.5001 | 0.4799 | 0.562 | 0.176 | 3.403 | 1.5166 | 0.2263 |
| | | ✓ | 0.822 | 0.7158 | 0.4498 | 0.5792 | 0.5838 | 0.03 | 5.4449 | 1.6876 | 0.0077 |
| ResNet50 | ✗ | ✗ | 0.2505 | 0.3762 | 0.8304 | 0.293 | 0.4128 | 0.6126 | **1.5653** | 1.0092 | 0.5302 |
| | | ✓ | 0.1794 | 0.3437 | 0.8755 | **0.2415** | 0.3658 | 0.6232 | 3.5983 | 1.5014 | 0.2312 |
| | ✓ | ✗ | 0.8538 | 0.7464 | 0.4149 | 0.675 | 0.651 | 0.0254 | 5.2797 | 1.8285 | -0.0783 |
| | | ✓ | 1.493 | 1.0126 | -0.0197 | 0.7632 | 0.6722 | -0.0304 | 4.1046 | 1.5906 | -0.0468 |
| ResNet101 | ✗ | ✗ | 0.4176 | 0.5144 | 0.7305 | 0.2604 | 0.3752 | 0.6178 | 2.6823 | 1.3048 | 0.5322 |
| | | ✓ | 0.187 | 0.3142 | 0.8705 | 0.4256 | 0.4719 | 0.445 | 2.3417 | 1.1524 | 0.5011 |
| | ✓ | ✗ | 0.814 | 0.714 | 0.4785 | 0.5543 | 0.5996 | -0.0976 | 5.3268 | 1.8533 | -0.2803 |
| | | ✓ | 1.5549 | 0.9883 | 0.0083 | 0.6594 | 0.638 | -0.0173 | 5.4934 | 1.7993 | -0.0701 |
| EfficientNet B0 | ✗ | ✗ | 0.2011 | 0.3496 | 0.8456 | 0.6401 | 0.6285 | 0.0434 | 2.6196 | 1.3196 | 0.4048 |
| | | ✓ | **0.1708** | 0.2938 | 0.8993 | 0.2454 | 0.3308 | 0.5745 | 1.8188 | 1.0976 | 0.5138 |
| | ✓ | ✗ | 0.3894 | 0.4788 | 0.7296 | 0.3304 | 0.4154 | 0.4635 | 3.5415 | 1.4078 | 0.1053 |
| | | ✓ | 0.3426 | 0.4649 | 0.7481 | 0.4368 | 0.5052 | 0.1245 | 3.9994 | 1.5346 | 0.0272 |
| EfficientNet B4 | ✗ | ✗ | 0.3442 | 0.4154 | 0.7846 | 0.5621 | 0.5656 | 0.1992 | 1.8014 | 1.0493 | 0.5446 |
| | ✓ | ✗ | 0.4325 | 0.4983 | 0.7253 | 0.4971 | 0.5244 | 0.3107 | 3.898 | 1.5122 | 0.1803 |
| EfficientNet B7 | ✗ | ✗ | 0.7521 | 0.6948 | 0.5255 | 0.7164 | 0.5867 | 0.1501 | 3.9136 | 1.5118 | 0.1067 |
| DINOv2 ViT-S-Reg | ✗ | ✗ | 1.4764 | 0.9438 | -0.124 | 0.6528 | 0.6068 | -0.0254 | 3.1641 | 1.4011 | 0.0026 |
| | | ✓ | 1.5061 | 0.9874 | 0.1093 | 0.7299 | 0.6455 | -0.1075 | 4.0756 | 1.5734 | -0.1107 |
| | ✓ | ✗ | 0.678 | 0.669 | 0.5249 | 0.6275 | 0.5999 | 0.09 | 4.8728 | 1.7811 | 0.0198 |
| | | ✓ | 1.6571 | 1.041 | 0.0092 | 0.7677 | 0.646 | -0.0293 | 4.1536 | 1.5651 | 0.001 |
| DINOv2 ViT-B-Reg | ✗ | ✗ | 1.3613 | 0.9498 | 0.1937 | 0.7688 | 0.6131 | -0.0051 | 6.3468 | 1.9548 | -0.1135 |
| | ✓ | ✗ | 0.4584 | 0.5341 | 0.6644 | 0.5802 | 0.563 | 0.2984 | 3.6433 | 1.51 | 0.1284 |
| DINOv2 ViT-L-Reg | ✗ | ✗ | 1.2595 | 0.9129 | 0.125 | 0.9686 | 0.721 | -0.0051 | 4.844 | 1.7005 | -0.0448 |

Table 5-5. Assessment of Self-supervised Pretraining under Feature Pyramid Network for MRD1, MRD2, and LF tasks.